\documentclass[a4paper,fleqn]{cas-dc}

\usepackage[numbers]{natbib}
\setcitestyle{numbers,square}

\let\CASoriginalbibliographystyle\bibliographystyle
\renewcommand{\bibliographystyle}[1]{%
  \CASoriginalbibliographystyle{unsrtnat}%
}

\usepackage{booktabs,array}

\def\tsc#1{\csdef{#1}{\textsc{\lowercase{#1}}\xspace}}
\tsc{WGM}
\tsc{QE}
\tsc{EP}
\tsc{PMS}
\tsc{BEC}
\tsc{DE}

\begin{document}
\let\WriteBookmarks\relax
\def\floatpagepagefraction{1}
\def\textpagefraction{.001}
\shorttitle{GFMR for RS OVSS}
\shortauthors{Jianzheng Wang et~al.}

\title [mode = title]{A Geometric-Transformation Feature-Adaptive Manifold Restoration Method for Open-Vocabulary Semantic Segmentation of Remote Sensing Images}                      



\author[1]{Jianzheng {Wang}}

\author[1]{Huan {Ni}}
\cormark[1]
\affiliation[1]{organization={ School of Remote Sensing and
Geomatics Engineering, Nanjing University of Information Science and Technology, China},
                city={ Nanjing},
                postcode={210044}, 
                country={China}}

\author[2]{Xiaonan {Niu}}
\affiliation[2]{organization={Nanjing Center, China Geological Survey},
                city={Nanjing},
                postcode={210016}, 
                country={China}}

\author[3]{Danfeng {Hong}}
\affiliation[3]{organization={School of Automation, Southeast University},
                city={Nanjing},
                postcode={210096}, 
                country={China}}

\author[1]{Haiyan {Guan}}

\cortext[cor1]{Corresponding author}


\begin{abstract}
The semantic information of objects in remote sensing images is typically invariant to geometric transformations from the dihedral group \(\mathrm{D}_4\). However, SAM3-based open-vocabulary semantic segmentation (OVSS) methods often exhibit inconsistent responses to different geometric transformations. To exploit this property and improve the stability of OVSS for remote sensing images, we propose a feature-adaptive manifold repair method based on dihedral-group geometric transformations. First, we introduce multi-scale harmonic-guided \(\mathrm{D}_4\) view selection (MH-D4VS) to select complementary candidate views from a set of geometrically transformed views. Next, we propose original-view-anchored adaptive manifold repair (OAMR), which uses the original view as an anchor and reliable cross-view information to selectively repair locally unreliable visual features. Finally, we develop pixel decoder test-time adaptation (PD-TTA) for SAM3, which fine-tunes only the parameters of the GroupNorm layers online during inference, thereby enhancing the model’s ability to adapt to sample-level distribution shifts. Experimental results show that the proposed method achieves an average mIoU of 55.6\% across eight remote sensing semantic segmentation benchmarks and delivers consistent performance improvements under different SAM3-based inference frameworks.

%
\end{abstract}


\maketitle

\section{Introduction}
\label{sec:introduction}
With the rapid development of Vision--Language Models (VLMs) and Vision Foundation Models (VFMs), Open-Vocabulary Semantic Segmentation (OVSS) has gradually emerged as a major paradigm for overcoming the constraints of fixed label spaces. Existing OVSS methods can be broadly categorized into training-based and training-free approaches. The current mainstream paradigm is still primarily built upon CLIP \cite{radford_learning_2021} and its adaptations for dense prediction. These methods mainly rely on reconstructing the internal attention maps or features of CLIP, as exemplified by MaskCLIP \cite{zhou_extract_2022}, SCLIP \cite{wang_sclip_2025}, ClearCLIP \cite{lan_clearclip_2025}, and ResCLIP \cite{yang_resclip_2024}. However, the CLIP image encoder is biased toward relatively large objects, while its text encoder is sensitive to the ordering of objects, as discussed in \cite{abbasi_clip_2025}. In fine-grained, multi-object remote sensing scenes, these biases may further exacerbate insufficient responses to local objects. Consequently, CLIP-based approaches still struggle to fully satisfy the requirements of remote sensing applications for precise spatial localization.

To address this limitation, another line of research combines CLIP with VFMs to exploit the finer-grained spatial structure provided by models such as DINO \cite{caron_emerging_2021,oquab_dinov2_2024} and SAM \cite{kirillov_segment_2023,ravi_sam_2025}. Representative methods include ProxyCLIP \cite{lan_proxyclip_2024}, CorrCLIP \cite{zhang_corrclip_2025}, and SegEarth-OV \cite{li_segearth-ov_2025}. Although such multi-model collaboration can jointly preserve semantic openness and structural resolution, the feature spaces, resolutions, and response mechanisms of different models are not inherently consistent. Therefore, additional feature alignment, information injection, or post-processing procedures are typically required. The introduction of SAM3 \cite{carion_sam_2025} provides new possibilities for constructing a more unified OVSS framework. Unlike the aforementioned approaches, SAM3 is designed for promptable concept segmentation and simultaneously performs concept recognition and pixel-level prediction within a unified query-based Transformer architecture. Nevertheless, a unified model architecture does not necessarily imply that its visual representations inherently conform to the geometric transformation relationships found in remote sensing scenes. Owing to the nadir-view imaging configuration of remote sensing imagery, the semantics of the same ground object generally remain invariant under geometric transformations from the dihedral group \(\mathrm{D}_4\). However, the SAM3 visual encoder produces different responses because such transformations alter spatial sampling, local context, and position-dependent representations. This study seeks to exploit these response discrepancies to repair and enhance visual representations, thereby improving segmentation performance.

However, simply enumerating all \(\mathrm{D}_4\) geometric transformations and directly averaging their predictions or features cannot guarantee the effective utilization of image information. To address this issue, we propose Multi-Scale Harmonic-Guided \(\mathrm{D}_4\) View Selection (MH-D4VS), which selects structurally informative and geometrically complementary transformations. Specifically, MH-D4VS constructs a compact complex harmonic orientation descriptor from multi-scale oriented edges and supplements it with chirality information to distinguish between rotations and reflections. It then evaluates the structural preferences of different geometric views according to the analytical action of \(\mathrm{D}_4\) transformations in the harmonic domain. By jointly considering structural quality, geometric diversity, and the balance among transformation families, MH-D4VS adaptively selects three complementary views from these non-identity candidates.

Building upon MH-D4VS, we further propose Original-View-Anchored Adaptive Manifold Repair (OAMR). Although the candidate views selected by MH-D4VS provide complementary geometric information, visual features extracted under different \(\mathrm{D}_4\) transformations exhibit varying degrees of response shifts. Consequently, candidate features at certain spatial locations may contain unstable or even erroneous semantic activations. Simple multi-view averaging or direct feature fusion may propagate these unreliable responses into the final representation, preventing the model from fully exploiting the complementary advantages of multi-view observations. Based on this observation, OAMR treats the original view as both the spatial and representation-state anchor. It evaluates feature consistency across different views within a unified coordinate system and establishes local reference structures using stable regions. Selective repair is performed when the feature from the original view deviates significantly from the local reference structure and the candidate views provide representations that are more consistent across views. In this manner, auxiliary views are used to correct locally unreliable responses while preserving the stable information already contained in the original-view representation.

After the multi-view geometric information has been selected and repaired, the semantic responses of the model to the current test sample may still be affected by cross-domain distribution shifts. This issue is particularly prominent in remote sensing imagery, where different regions often exhibit substantial variations in imaging conditions, object scales, texture distributions, and scene compositions. As a result, the fixed feature distributions learned by a pretrained model may not remain consistently reliable across different scenes. To address this problem, we design Pixel Decoder Test-Time Adaptation for SAM3 (PD-TTA), a test-time adaptation strategy tailored to the SAM3 dense visual decoder. PD-TTA uses the predictive uncertainty of the current sample as a self-supervised signal and performs sample-specific distribution recalibration by imposing an entropy-based constraint on the decoder representations. This enables the model to dynamically adjust its semantic responses according to the current input and improves prediction stability under varying scene conditions. Adaptation is performed independently for each test sample without requiring additional annotated data. During online adaptation, only the parameters of the GroupNorm layers are fine-tuned, while all other model parameters remain frozen. PD-TTA therefore improves the stability of semantic predictions without modifying the main inference architecture of SAM3.

The main contributions of this work are summarized as follows:

\begin{enumerate}
    \item We propose MH-D4VS, which characterizes directional structures using multi-scale complex harmonics and chirality. By jointly considering structural preference, \(\mathrm{D}_4\) geometric diversity, and transformation-family balance, MH-D4VS adaptively selects complementary non-identity views, thereby improving the geometric stability of the feature representations produced by the pretrained SAM3 model.

    \item We propose OAMR, which imposes reliability constraints on multi-view visual features through original-view-anchored local manifold modeling. OAMR constructs a local manifold reference using an adaptive local subspace and evaluates the manifold consistency of the original and candidate features through refined cross-view metrics. It repairs locations whose features deviate significantly from the local manifold and are supported by reliable candidate views. In addition, Cross-Scale Guidance is introduced to provide complementary fine-grained spatial evidence.

    \item We design PD-TTA for the dense decoding architecture of SAM3. By integrating MH-D4VS, OAMR, and PD-TTA, we establish a new OVSS inference framework for remote sensing imagery, termed Geometry-Transformation-Aware Feature-Adaptive Manifold Restoration (GFMR). Experiments on remote sensing datasets demonstrate that GFMR achieves higher segmentation accuracy than existing OVSS methods.
\end{enumerate}

\section{Related Work}
\label{sec:related_work}

\subsection{Training-Based OVSS}
\label{subsec:training_based_ovss}

Training-based OVSS methods are typically built upon large-scale pretrained Vision--Language Models (VLMs) and further adapted to dense prediction tasks using data with pixel-level or region-level annotations. According to how visual representations are integrated with class semantics, existing methods can be broadly divided into two categories: mask-classification-based methods and dense-feature-adaptation-based methods.

Mask-classification-based methods generally first generate class-agnostic regions or masks and then perform open-vocabulary category recognition using visual--language similarity. OpenSeg \cite{ghiasi_scaling_2021}, ZegFormer \cite{ding_decoupling_2022}, OVSeg \cite{liang_open-vocabulary_2023}, and MasQCLIP \cite{xu_masqclip_2023} follow this paradigm. By decoupling region generation from open-vocabulary category recognition, these methods extend the language semantics acquired during pretraining to previously unseen categories. Another line of research directly adapts the dense representations of visual--language models. LSeg \cite{li_language-driven_2022} constructs an open-vocabulary prediction space through pixel-level visual--language alignment. SAN \cite{xu_side_2023}, SED \cite{xie_sed_2023}, and CAT-Seg \cite{cho_cat-seg_2023} further improve the spatial discriminability of visual--language features using side networks, adaptation modules, or cost-aggregation mechanisms. SegCLIP \cite{luo_segclip_2022} and GroupViT \cite{xu_groupvit_2022} learn region grouping or visual-token aggregation from image--text supervision, providing alternative approaches to open-vocabulary region modeling.

Overall, training-based methods can explicitly reduce the representation gap between image-level pretraining and pixel-level prediction by leveraging task-relevant data, and therefore generally exhibit strong adaptability to downstream tasks. However, their performance depends on additional training data, optimization procedures, and computational resources. Consequently, their deployment flexibility is limited in applications involving rapidly changing category spaces or scarce annotations.

\subsection{Training-Free OVSS}
\label{subsec:training_free_ovss}

Training-free OVSS aims to perform open-vocabulary dense prediction by directly exploiting the visual and semantic knowledge already encoded in pretrained models, without optimizing model parameters on the target data. Existing studies have primarily explored three directions: extracting spatial information from the internal representations of CLIP, exploiting structural priors provided by Visual Foundation Models (VFMs), and combining VLMs with VFMs to jointly preserve open-vocabulary semantics and fine-grained spatial representations. These three research directions have ultimately begun to converge toward unified foundation-model frameworks.

\subsubsection{CLIP Internal Representation Adaptation}
\label{subsubsec:clip_internal_adaptation}

Methods in this category directly exploit the patch representations and attention relationships already encoded within CLIP, enabling a visual encoder originally designed for image-level recognition to produce spatial responses that are more suitable for pixel-level prediction. MaskCLIP \cite{zhou_extract_2022} directly recovers patch-level visual representations from CLIP. SCLIP \cite{wang_sclip_2025} and ClearCLIP \cite{lan_clearclip_2025} improve spatial consistency by modifying token interactions within the self-attention mechanism. NACLIP \cite{hajimiri_pay_2024} further strengthens local neighborhood relationships to improve localization capability. ResCLIP \cite{yang_resclip_2024}, in contrast, exploits the spatial correlations retained in intermediate layers to enhance high-level pixel-wise dense representations.

These studies demonstrate that the internal representations of CLIP retain local information that can be used for pixel-level prediction. Nevertheless, the CLIP pretraining objective primarily focuses on global image--text alignment, while local boundaries, spatial adjacency relationships, and fine-grained object structures are not directly optimized. Therefore, relying exclusively on the reconstruction of internal attention maps or patch representations remains insufficient to fully bridge the gap between image-level visual--language pretraining and fine-grained spatial understanding.

\subsubsection{Visual Foundation Models for Structure Discovery}
\label{subsubsec:vfm_structure_discovery}

Visual foundation models generally provide more explicit local correspondences and category-agnostic spatial structures than visual--language models. The DINO family \cite{caron_emerging_2021,oquab_dinov2_2024} learns visual representations with strong local correspondence capabilities through large-scale self-supervised learning. Building upon such representations, TokenCut \cite{wang_tokencut_2023} and CutLER \cite{wang_cut_2023,wang_videocutler_2024} further discover salient regions and instance structures from self-supervised features. The SAM family \cite{kirillov_segment_2023,ravi_sam_2025}, meanwhile, exhibits strong category-agnostic region segmentation capabilities. Based on this capability, E-SAM \cite{zhang_e-sam_2025} constructs a training-free entity segmentation framework that addresses the over-segmentation and under-segmentation problems that may occur during automatic mask generation with SAM. OmniSAM \cite{zhong_omnisam_2025} further explores the ability of SAM to model cross-field-of-view information in panoramic scenes.

These methods can effectively characterize the regional organization and object structures within an image. However, their visual embeddings are generally not directly aligned with an open-vocabulary textual semantic space. Consequently, VFMs are better suited to serving as sources of spatial structure and have difficulty independently performing OVSS.

\subsubsection{VLM--VFM Collaborative OVSS}
\label{subsubsec:vlm_vfm_collaboration}

To jointly exploit the open-vocabulary semantic capabilities of VLMs and the fine-grained spatial structures of VFMs, multi-model collaboration has gradually become an important direction in OVSS. Some studies enhance the visual representations of arbitrary categories using external visual references. For example, ReCo \cite{shin_reco_2022}, OVDiff \cite{karazija_diffusion_2025}, FOSSIL \cite{barsellotti_fossil_2024}, and FreeDA \cite{barsellotti_training-free_2024} employ retrieval systems, generative models, or offline visual references to construct richer visual support information for textual categories.

Another category of methods directly establishes representational connections between visual--language models and self-supervised visual models. Talk2DINO \cite{barsellotti_talking_2025} maps the textual semantics of CLIP to the patch-level visual features of DINO-v2, whereas CLIP-DINOiser \cite{wysoczanska_clip-dinoiser_2025} uses localization priors provided by self-supervised features to improve the local representations of CLIP. Because these approaches require additional mapping modules or lightweight feature learning, they differ to some extent from a strictly training-free setting with no parameter updates.

In collaborative frameworks that strictly require no additional training, a more common strategy is to use external VFMs to provide spatial structural guidance. ProxyCLIP \cite{lan_proxyclip_2024} and CorrCLIP \cite{zhang_corrclip_2025} improve the spatial responses of CLIP by exploiting structural relationships derived from external visual models. CASS \cite{kim_distilling_2025} further introduces object-level contextual information. Trident \cite{shi_harnessing_2024} jointly employs foundation models such as CLIP, DINO, and SAM to enhance OVSS ability through their complementary representations.

Overall, VLM--VFM collaboration can simultaneously preserve open-vocabulary semantics and spatial structure. However, such approaches generally depend on multiple foundation models with different pretraining objectives and feature spaces, requiring feature mapping, attention guidance, or cross-model information interaction to establish connections among them.

\subsubsection{From Heterogeneous Collaboration to Unified Foundation Models}
\label{subsubsec:unified_foundation_models}

In contrast to the aforementioned heterogeneous model collaboration paradigms, the introduction of SAM3 \cite{carion_sam_2025} provides a new foundation for OVSS frameworks. SAM3 is designed for promptable concept segmentation and jointly models concept recognition and dense prediction within a unified architecture, allowing open-vocabulary semantic understanding and spatial segmentation to share the same visual representation system. Based on this framework, ActiveSAM \cite{tien_activesam_2026} further improves inference efficiency by dynamically selecting active categories according to the input image, thereby reducing redundant computation during large-scale open-vocabulary inference.

To some extent, SAM3 reduces the cross-model interfaces and representational discrepancies associated with conventional VLM--VFM collaboration. Nevertheless, a unified architecture does not necessarily imply that its visual representations can naturally adapt to the spatial properties of different data domains. For remote sensing imagery, arbitrary object orientations caused by nadir-view imaging, substantial scale variations, and densely distributed objects continue to pose new challenges to spatial representation and geometric adaptability.

\subsection{OVSS for Remote Sensing Images}
\label{subsec:remote_sensing_ovss}

Extending OVSS to remote sensing imagery is not merely a matter of transferring a model to a new data domain. Nadir-view imaging means that objects belonging to the same category generally lack a fixed orientation. Variations in spatial resolution and object size also introduce substantial scale differences, while high-resolution scenes often contain densely distributed small objects, complex boundaries, and large continuous regions. Therefore, in addition to preserving semantic generalization to open categories, remote sensing OVSS must explicitly address challenges such as orientation and scale variations.

Existing studies have explored spatial-detail recovery and domain adaptation \cite{chen_category-certainty_2022, chen_memory-contrastive_2023} for remote sensing imagery from different perspectives. SegEarth-OV \cite{li_segearth-ov_2025} restores high-resolution visual--language representations through feature upsampling. A related annotation-free extension \cite{li_annotation-free_2025} further introduces AlignEarth, extending open-vocabulary segmentation to the Synthetic Aperture Radar (SAR) modality. GSNet \cite{ye_towards_2024} employs a dual-branch architecture to combine remote-sensing-specific priors with general visual--language semantics, whereas SkySense-O \cite{zhu_skysense-o_2025} enhances open-world visual representations through remote-sensing-specific visual--language pretraining. ConInfer \cite{chen_coninfer_2026} further considers semantic dependencies among different spatial units in large-scale remote sensing scenes and improves regional prediction consistency through contextual joint inference.

The strong category-agnostic spatial modeling capability of the SAM family has also been increasingly introduced into open-world remote sensing applications. AerOSeg \cite{dutta_aeroseg_2025-1} and SCORE \cite{huang_score_2025} use SAM features to provide structural guidance and regional context. RemoteSAM \cite{yao_remotesam_2025} and InstructSAM \cite{zheng_instructsam_2025} utilize SAM-based pipelines to unify segmentation with broader interpretation tasks. Following the introduction of SAM3, SegEarth-OV3 \cite{li_segearth-ov3_2025} enhances its application to training-free remote sensing OVSS through prediction fusion and class-presence filtering. ProC-SAM3 \cite{song_prompt-calibrated_2026}, meanwhile, improves remote sensing OVSS inference through prompt expansion and reliable-response aggregation.

Overall, existing remote sensing OVSS methods have made continuous progress in high-resolution feature extraction, domain adaptation, cross-model collaboration, contextual modeling, and the utilization of SAM-based foundation models. Nevertheless, existing SAM3-based remote sensing OVSS methods primarily focus on inference mechanisms such as prediction fusion, category filtering, and prompt calibration, while the multi-view geometric information arising from rotations and reflections in remote sensing imagery has yet to be systematically exploited.

On the one hand, the structural information provided by different geometric views is not equivalent for a given scene, and simply enumerating all transformations can introduce redundant observations. On the other hand, even after different views are realigned to a unified coordinate system, the high-dimensional features at corresponding spatial locations may still exhibit substantial response discrepancies. Direct averaging or indiscriminate fusion may therefore propagate unreliable responses. Consequently, identifying representative and complementary geometric information from the multi-view observations generated by a unified foundation model itself, and reliably exploiting this information to enhance the original visual representation, remains an important research problem in OVSS for remote sensing imagery.

\section{Methodology}
\label{sec:methodology}

\subsection{Overall Framework}
\label{subsec:overall_framework}

\begin{figure*}
	\centering
	\includegraphics[width=.9\textwidth]{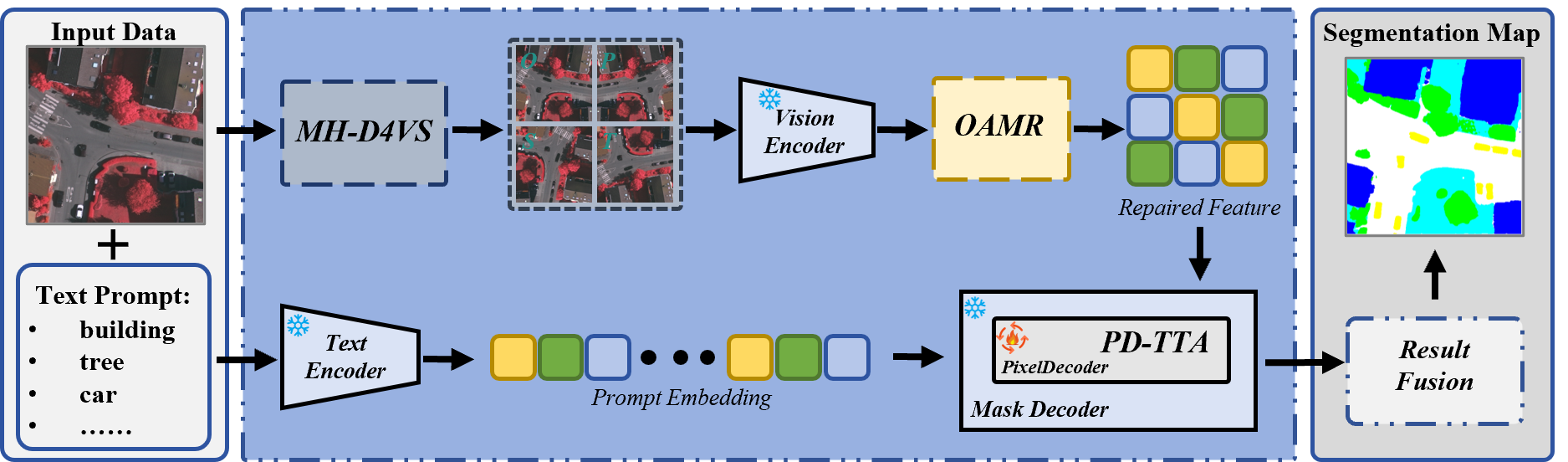}
	\caption{Overview of the proposed Geometry-Transformation-Aware Feature-Adaptive Manifold Restoration framework (GFMR). GFMR consists of three components including MH-D4VS, OAMR and PD-TTA.}
	\label{fig:framework}
\end{figure*}

The overall pipeline of the proposed GFMR framework for OVSS of remote sensing images is illustrated in Figure~\ref{fig:framework}. GFMR consists of three interconnected components. First, MH-D4VS analyzes the multi-scale directional structures of the input image and selects three candidate views with both structural representativeness and geometric complementarity from the seven non-identity \(\mathrm{D}_4\) transformations. Subsequently, the features from the selected candidate views are realigned to the original coordinate system. Taking the original view as the anchor, OAMR constructs a local manifold reference and combines adaptive subspace modeling with cross-view consistency measurements to identify locally anomalous features that deviate from the manifold. Selective repair is performed when the candidate views provide more reliable representations. Finally, PD-TTA uses the semantic uncertainty of the current test image to perform sample-specific adaptation of the affine parameters of the GroupNorm layers online in the PixelDecoder that participate in the forward pass. 

\subsection{Multi-Scale Harmonic-Guided \texorpdfstring{\(\mathrm{D}_4\)}{D4} View Selection}
\label{subsec:mh_d4vs}

\begin{figure*}
	\centering
	\includegraphics[width=1.0\textwidth]{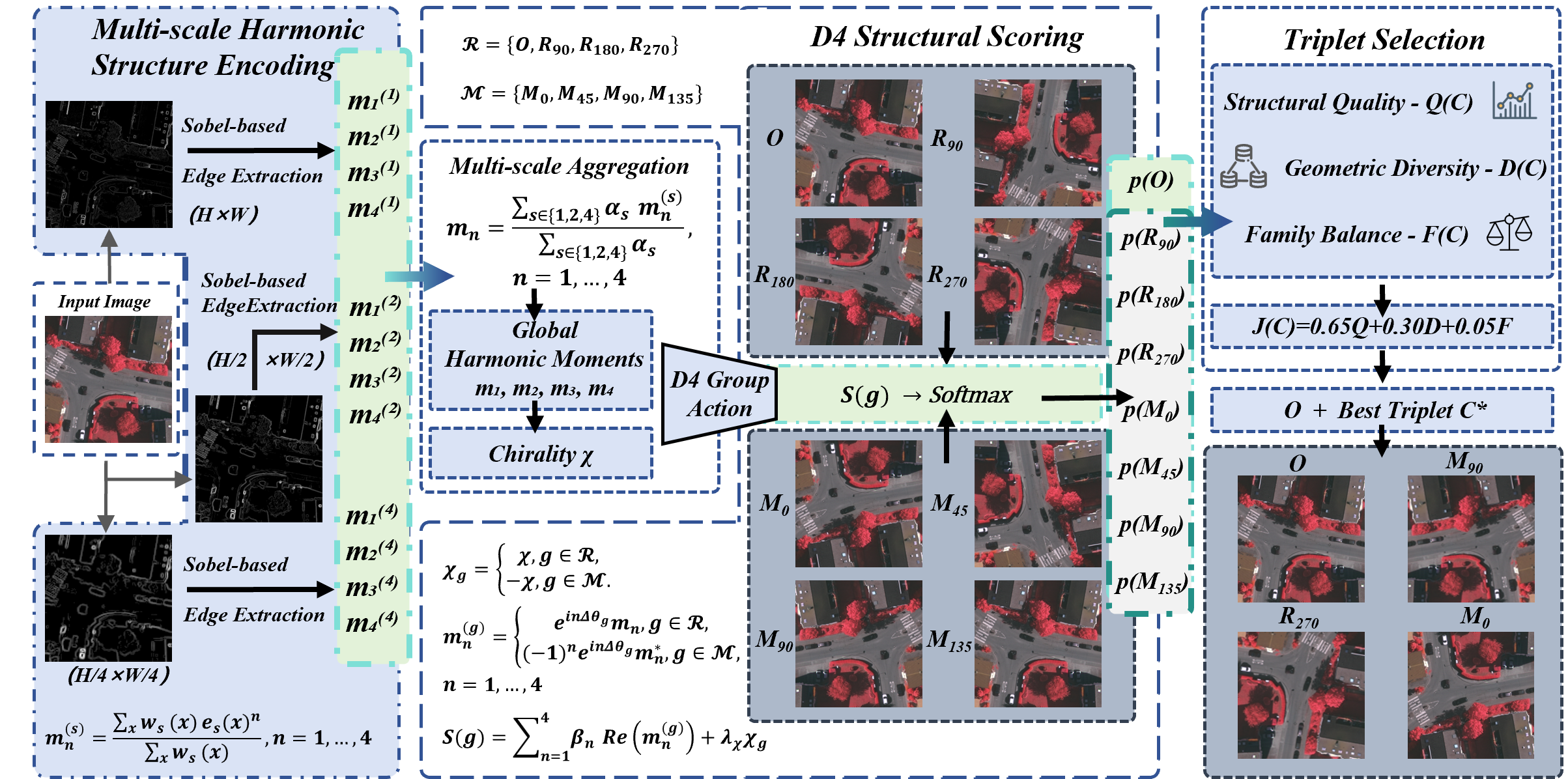}
	\caption{Overview of MH-D4VS. Multi-scale oriented edges are first used to construct and aggregate first- to fourth-order complex harmonic moments, while chirality is computed to capture handedness information. The action of the \(\mathrm{D}_4\) group is then applied to derive the structural preferences of different geometric views. Finally, the optimal three-view combination \(C^{\ast}\) is selected by jointly considering structural quality \(Q(C)\), geometric diversity \(D(C)\), and transformation-family balance \(F(C)\). The selected views are combined with the original view \(O\) to form the multi-view representation used in subsequent processing.}
	\label{fig:mhd4vs}
\end{figure*}

Roads, building boundaries, and other artificial structures in remote sensing images often exhibit prominent directional patterns. Different rotated and reflected views alter the relative relationships between these structures and their local contexts, thereby producing diverse visual responses. However, the structural contributions of different \(\mathrm{D}_4\) transformations are not equivalent for a given scene. Directly enumerating all transformations not only introduces redundant observations but also increases the computational cost of subsequent multi-view feature extraction. To address this issue, we propose the Multi-Scale Harmonic-Guided \(\mathrm{D}_4\) View Selection (MH-D4VS) module as shown in figure~\ref{fig:mhd4vs}. Before visual encoding, MH-D4VS selects three representative and geometrically complementary candidate views from the seven non-identity \(\mathrm{D}_4\) transformations according to the directional structures of the input image.

MH-D4VS uses the directional structures of the input image as the basis for view selection. It characterizes the relative preferences of different geometric views through image-domain structural responses and further constrains the geometric complementarity among candidates at the set level. Specifically, the module first describes the directional structures of the input image using multi-scale complex harmonics and introduces chirality to capture the handedness differences between rotations and reflections. It then derives the structural response of each transformation through the analytical action of the \(\mathrm{D}_4\) group in the harmonic domain. Finally, the optimal three-view combination is determined by jointly considering structural quality, geometric diversity, and transformation-family balance.

\subsubsection{Multi-Scale Harmonic Structural Encoding}
\label{subsubsec:multiscale_harmonic_encoding}

Given an input image \(I\), we first convert it into a grayscale representation and construct a multi-scale structural description over the scale set
\(\mathcal{S}=\{1,2,4\}\). At each scale \(s\), the horizontal and vertical gradients, denoted by \(G_x^{(s)}(x)\) and \(G_y^{(s)}(x)\), respectively, are computed using the horizontal and vertical Sobel operators. The gradient magnitude is then defined as
\begin{equation}
G^{(s)}(x)
=
\sqrt{
    \left(G_x^{(s)}(x)\right)^2
    +
    \left(G_y^{(s)}(x)\right)^2
}.
\label{eq:gradient_magnitude}
\end{equation}

To reduce interference from weak textures and noise in the global orientation statistics, only locations with sufficiently strong gradient responses are retained for subsequent computation. Let \(M_s(x)\in\{0,1\}\) denote the strong-edge indicator at scale \(s\). The corresponding edge weight is defined as
\begin{equation}
\left\{
\begin{aligned}
w_s(x) &= G^{(s)}(x)M_s(x), \\
M_s(x) &\in \{0,1\}.
\end{aligned}
\right.
\label{eq:edge_weight}
\end{equation}
Here, \(M_s(x)=1\) indicates that location \(x\) is retained for structural statistics, whereas \(M_s(x)=0\) indicates that it is discarded. By preserving variations in gradient magnitude, \(w_s(x)\) assigns greater contributions to more prominent edges in the overall orientation description.

Based on this representation, the two-dimensional edge orientation is expressed as a complex-valued oriented tangent:
\begin{equation}
e_s(x)
=
\frac{
    -G_y^{(s)}(x)
    +
    \mathrm{i}G_x^{(s)}(x)
}{
    \max\left(G^{(s)}(x),\varepsilon\right)
},
\label{eq:oriented_tangent}
\end{equation}
where \(\mathrm{i}=\sqrt{-1}\), and \(\varepsilon>0\) is a sufficiently small constant introduced for numerical stability. At locations with nonzero gradients, \(e_s(x)\) is approximately a unit complex number and can be written as
\begin{equation}
e_s(x)\approx \exp\left(\mathrm{i}\theta_s(x)\right),
\end{equation}
where \(\theta_s(x)\) denotes the oriented edge-tangent direction at location \(x\). Unlike representations that describe only an unoriented principal axis, this oriented complex representation preserves directional sign information. Consequently, odd-order harmonics can further distinguish structures exhibiting opposite directional relationships.

Using the above representation, the \(n\)-th-order weighted complex harmonic moment at scale \(s\) is computed as
\begin{equation}
m_n^{(s)}
=
\frac{
    \displaystyle\sum_x w_s(x)e_s(x)^n
}{
    \displaystyle\sum_x w_s(x)
},
\qquad
n=1,\ldots,4.
\label{eq:scale_harmonic_moment}
\end{equation}
Here, \(m_n^{(s)}\) represents the \(n\)-th-order directional-structure statistic at scale \(s\). Different harmonic orders correspond to different directional periodicities. The first-order (\(n=1\)) and third-order (\(n=3\)) harmonics preserve asymmetric information about oriented structures, while the second-order harmonic (\(n=2\)) primarily characterizes axial structures. The fourth-order harmonic (\(n=4\)) responds to scene patterns exhibiting four-fold directional characteristics. Therefore, the first- to fourth-order harmonics describe the directional organization of remote sensing images from complementary perspectives without explicitly estimating a single dominant orientation.

Because different scales contain different amounts of informative structural content, we further perform adaptive aggregation according to the edge-response strength at each scale. Let \(X_s\) denote the set of all spatial locations at scale \(s\). The scale weight is defined as
\begin{equation}
\alpha_s
=
\frac{1}{\lvert X_s\rvert}
\sum_{x\in X_s} w_s(x),
\label{eq:scale_weight}
\end{equation}
where \(\lvert X_s\rvert\) is the number of spatial locations at scale \(s\), and \(\alpha_s\) reflects the overall strength of the informative edge responses at that scale. Accordingly, the global \(n\)-th-order complex harmonic moment is defined as
\begin{equation}
m_n
=
\frac{
    \displaystyle\sum_{s\in\mathcal{S}}
    \alpha_s m_n^{(s)}
}{
    \displaystyle\sum_{s\in\mathcal{S}}\alpha_s
},
\qquad
n=1,\ldots,4.
\label{eq:global_harmonic_moment}
\end{equation}

This aggregation is performed only among moments of the same harmonic order, thereby preserving the directional interpretation associated with each order. Meanwhile, scales with stronger informative edge responses receive larger weights, allowing the representation to capture both local contours and broader directional structures. The resulting global multi-scale harmonic representation is \(\boldsymbol{m} = \left\{m_1,m_2,m_3,m_4\right\}\), which serves as the basis for the subsequent construction of chirality and the modeling of structural responses under different \(\mathrm{D}_4\) geometric views.

\subsubsection{Cross-Order Harmonic Chirality Representation}
\label{subsubsec:cross_order_chirality}

Relying solely on individual harmonic responses may be insufficient to distinguish rotated and reflected views with similar orientation statistics. To capture the changes in handedness induced by reflection, we construct a cross-order chirality representation after multi-scale aggregation. Specifically, we define \(I_{12}=m_1^2m_2^{\ast}, I_{13}=m_1^3m_3^{\ast}, I_{24}=m_2^2m_4^{\ast}\). Here, \((\cdot)^{\ast}\) denotes complex conjugation, and \(I_{12}\), \(I_{13}\), and \(I_{24}\) are cross-order complex statistics obtained by coupling harmonics of different orders. The final normalized chirality is defined as
\begin{equation}
\chi=\frac{\operatorname{Im}(I_{12})+\operatorname{Im}(I_{13})+\operatorname{Im}(I_{24})}{\lvert I_{12}\rvert+\lvert I_{13}\rvert+\lvert I_{24}\rvert+\varepsilon}.
\label{eq:normalized_chirality}
\end{equation}
Here, \(\operatorname{Im}(\cdot)\) denotes the imaginary part of a complex number, \(\lvert\cdot\rvert\) denotes its magnitude, and \(\varepsilon\) is a small positive constant introduced for numerical stability. The resulting chirality satisfies \(-1\leq\chi\leq 1\).

These cross-order combinations exploit the phase relationships among harmonics of different orders to construct a reflection-sensitive descriptor. Under rotation, the phase shifts associated with different harmonic orders cancel within the cross-order combinations. Under reflection, the complex-conjugate relationship reverses the signs of the imaginary parts of these combinations. Therefore, \(\chi\) compensates for the limited ability of harmonic real parts alone to distinguish rotations from reflections and provides handedness information for the subsequent structural scoring of \(\mathrm{D}_4\) views.

\subsubsection{Group-Aware Structural Response Modeling}
\label{subsubsec:group_aware_response}

After obtaining the global harmonic representation and chirality, MH-D4VS evaluates the structural responses of different geometric views using the analytical transformation rules of the \(\mathrm{D}_4\) group in the harmonic domain. We partition \(\mathrm{D}_4\) into a rotation subset and a reflection subset:
\begin{equation}
\left\{
\begin{aligned}
\mathcal{R}=&\{O,R_{90},R_{180},R_{270}\}, \\
\mathcal{M}=&\{M_0,M_{45},M_{90},M_{135}\}.
\end{aligned}
\right.
\label{eq:subsets}
\end{equation}
Here, \(O\) denotes the identity transformation, \(\mathcal{R}\) contains four rotation states, and \(\mathcal{M}\) contains four reflection states. Thus, \(\mathrm{D}_4=\mathcal{R}\cup\mathcal{M}\).

Since complex harmonics follow explicit transformation rules under rotation and reflection, the \(n\)-th-order harmonic moment after applying transformation \(g\) can be expressed as
\begin{equation}
m_n^{(g)}=\begin{cases}
e^{\mathrm{i}n\Delta\theta_g}m_n, & g\in\mathcal{R}, \\
(-1)^n e^{\mathrm{i}n\Delta\theta_g}m_n^{\ast}, & g\in\mathcal{M},
\end{cases}
\qquad n=1,\ldots,4.
\label{eq:group_harmonic_response}
\end{equation}
Here, \(m_n\) is the global \(n\)-th-order harmonic moment of the original view, and \(m_n^{(g)}\) denotes the corresponding harmonic response after transformation \(g\). The angle \(\Delta\theta_g\) is determined by the geometric action of the corresponding \(\mathrm{D}_4\) transformation and follows the rotation and reflection conventions adopted in this work. Furthermore, \(n\in\{1,2,3,4\}\) denotes the harmonic order, and \((\cdot)^{\ast}\) denotes complex conjugation. Rotations primarily induce harmonic phase shifts, whereas reflections additionally involve complex conjugation and a parity-dependent factor determined by the harmonic order. This analytical relationship allows the structural responses of different \(\mathrm{D}_4\) views to be derived directly from the original harmonic descriptor, without recomputing orientation statistics for each transformed image.

Correspondingly, chirality preserves its sign under rotation, whereas reflection reverses handedness. We therefore define
\begin{equation}
\chi_g=\begin{cases}
\chi, & g\in\mathcal{R}, \\
-\chi, & g\in\mathcal{M}.
\end{cases}
\label{eq:group_chirality_response}
\end{equation}
Here, \(\chi\) is the chirality of the original view obtained in the preceding subsection, and \(\chi_g\) denotes the chirality response associated with transformation \(g\). We then construct the structural score of view \(g\) using the harmonic responses and chirality:
\begin{equation}
S(g)=\sum_{n=1}^{4}\beta_n\,\operatorname{Re}\!\left(m_n^{(g)}\right)+\lambda_{\chi}\chi_g.
\label{eq:view_structural_score}
\end{equation}
Here, \(\operatorname{Re}(\cdot)\) denotes the real part of a complex number, \(\beta_n\) is the weight of the \(n\)-th-order harmonic and controls the relative contribution of the corresponding directional structure, and \(\lambda_{\chi}\) is the chirality weight. In this work, we set \((\beta_1,\beta_2,\beta_3,\beta_4)=(0.35,0.70,0.20,0.50), \lambda_{\chi}=0.20\).

Finally, a temperature-scaled softmax converts the structural scores of all \(\mathrm{D}_4\) elements into continuous structural preferences:
\begin{equation}
p(g)=\frac{\exp\!\left(S(g)/T\right)}{\sum_{h\in\mathrm{D}_4}\exp\!\left(S(h)/T\right)}.
\label{eq:structural_preference}
\end{equation}
Here, \(T\) is the temperature parameter, which is set to \(T=0.10\). The probability \(p(g)\) represents the relative structural preference of the input image within the \(\mathrm{D}_4\) transformation space. A larger \(p(g)\) indicates greater alignment between the corresponding geometric view and the directional structures of the current scene. This distribution characterizes the structural compatibility of different candidate views and provides a structural-quality criterion for the subsequent diversity-aware joint selection of three views.

\subsubsection{Diversity-Aware Triplet Selection}
\label{subsubsec:diversity_aware_selection}

The structural preference \(p(g)\) defined above describes the relative compatibility between each \(\mathrm{D}_4\) transformation and the directional structures of the current image. However, independently selecting the three non-identity transformations with the highest probabilities may yield candidates that are geometrically similar or even belong to the same transformation family, thereby reducing the complementarity of multi-view observations. We therefore jointly select the three views as a set, explicitly accounting for their geometric differences and transformation-family composition while preserving structural compatibility.

Let the set of non-identity candidates be
\begin{equation}
\mathcal{G}=\mathrm{D}_4\setminus\{O\}=\{R_{90},R_{180},R_{270},M_0,M_{45},M_{90},M_{135}\}.
\label{eq:nonidentity_candidates}
\end{equation}
Here, \(O\) denotes the original view. Considering all triplets \(C\subset\mathcal{G}\) with \(\lvert C\rvert=3\), the seven non-identity transformations yield \(\binom{7}{3}=35\) candidate combinations. We first define the structural-quality term as
\begin{equation}
Q(C)=\frac{\sum_{g\in C}p(g)}{\sum_{h\in\mathcal{G}}p(h)},
\label{eq:structural-quality}
\end{equation}
where \(p(g)\) is the structural probability of view \(g\) obtained in the preceding subsection. The term \(Q(C)\) measures the fraction of the total non-identity structural probability mass captured by triplet \(C\). A larger \(Q(C)\) indicates that the selected views are collectively more compatible with the directional structures of the input image.

Structural quality alone cannot prevent geometric redundancy among candidates. To address this issue, we represent each \(\mathrm{D}_4\) element \(g\) by its corresponding two-dimensional orthogonal transformation matrix \(A_g\). The normalized geometric distance between two transformations is defined as
\begin{equation}
d(g_i,g_j)=\frac{\lVert A_{g_i}-A_{g_j}\rVert_F}{\sqrt{8}},
\label{eq:geometric_distance}
\end{equation}
where \(\lVert\cdot\rVert_F\) denotes the Frobenius norm. The distance calculation includes both the three candidate views and the original view \(O\), thereby evaluating the overall coverage of the complete four-view set \(\mathcal{V}_C=\{O,g_1,g_2,g_3\}\) in the \(\mathrm{D}_4\) transformation space. Let \(d_{\mathrm{mean}}(C)\) and \(d_{\mathrm{min}}(C)\) denote the mean and minimum distances, respectively, over all pairs of distinct views in \(\mathcal{V}_C\). Geometric diversity is then defined as
\begin{equation}
D(C)=\frac{1}{2}d_{\mathrm{mean}}(C)+\frac{1}{2}d_{\mathrm{min}}(C).
\label{eq:geometric_diversity}
\end{equation}
The mean distance measures the overall geometric spread of the view set, while the minimum distance discourages any two views from being excessively similar. Together, these terms increase geometric coverage in \(\mathrm{D}_4\) while reducing redundant observations.

In addition to geometric distance, rotations and reflections provide different types of geometric observations. Let the non-identity rotation subset and the reflection subset be
\begin{equation}
\left\{
\begin{aligned}
\mathcal{R}'=&\{R_{90},R_{180},R_{270}\}, \\
\mathcal{M}=&\{M_0,M_{45},M_{90},M_{135}\}.
\end{aligned}
\right.
\label{eq:subsets-non}
\end{equation}

We accordingly define the transformation-family balance term as
\begin{equation}
F(C)=\begin{cases}
1, & C\cap\mathcal{R}'\neq\varnothing \;\land\; C\cap\mathcal{M}\neq\varnothing, \\
0, & \text{otherwise}.
\end{cases}
\label{eq:family_balance}
\end{equation}
This term encourages each candidate triplet to include both rotated and reflected views, preventing the selected observations from being concentrated entirely within a single transformation family. Since it primarily serves as a discrete constraint on the composition of the triplet, it is assigned a relatively small weight in the final objective.

Combining structural quality, geometric diversity, and transformation-family balance, we define the joint triplet score as
\begin{equation}
J(C)=\lambda_Q Q(C)+\lambda_D D(C)+\lambda_F F(C),
\label{eq:joint_triplet_score}
\end{equation}
where \(\lambda_Q\), \(\lambda_D\), and \(\lambda_F\) control the relative contributions of the three terms. In our experiments, these weights are set to \(\lambda_Q=0.65\), \(\lambda_D=0.30\), and \(\lambda_F=0.05\). The optimal candidate triplet is then obtained as
\begin{equation}
C^{\ast}=\operatorname*{arg\,max}_{C\subset\mathcal{G},\,\lvert C\rvert=3} J(C).
\label{eq:optimal_triplet}
\end{equation}
The larger weight assigned to structural quality ensures that candidate selection remains primarily driven by the directional structures of the current image. Geometric diversity further discourages transformations with high probabilities but strong mutual redundancy, while transformation-family balance promotes complementarity between rotations and reflections. Together, these three terms turn independent view ranking into joint optimization at the set level.

After obtaining the optimal candidate set \(C^{\ast}\), we sort its three elements in descending order of their structural probabilities \(p(g)\) and designate them as the \emph{Principal}, \emph{Secondary}, and \emph{Third} candidate views, respectively:
\begin{equation}
p(P)\geq p(S)\geq p(T), \qquad P,S,T\in C^{\ast}.
\label{eq:candidate_ordering}
\end{equation}

The views \(P\), \(S\), and \(T\) are therefore not selected directly as the global top three among all seven non-identity \(\mathrm{D}_4\) transformations according to \(p(g)\). Instead, \(J(C)\) first identifies the optimal triplet with overall complementarity, after which the candidates are ordered within this set. Consequently, a transformation with a relatively high structural probability may still be excluded if it is geometrically redundant with other candidates. The final multi-view set is expressed as
\begin{equation}
\mathcal{V}=\{O\}\cup C^{\ast}=\{O,P,S,T\}.
\label{eq:final_view_set}
\end{equation}
Here, \(O\) preserves the original spatial reference, while \(P\), \(S\), and \(T\) provide auxiliary observations jointly constrained by structural compatibility and geometric complementarity. These views serve as inputs to subsequent cross-view feature modeling and selective repair.

\subsection{Original-Anchored Adaptive Multi-View Manifold Repair (OAMR)}
\label{subsec:oamr}

Following MH-D4VS, we obtain a multi-view set consisting of the original view \(O\) and three complementary candidate views \(P\), \(S\), and \(T\). Although these views depict the same scene, their encoded visual features reside in different geometric coordinate systems. Even after spatial alignment, the high-dimensional features at corresponding locations may differ in reliability. Direct cross-view averaging or indiscriminate fusion may propagate locally anomalous responses into the final representation and compromise the stable semantic information already present in the original view.

\begin{figure*}
	\centering
	\includegraphics[width=1.0\textwidth]{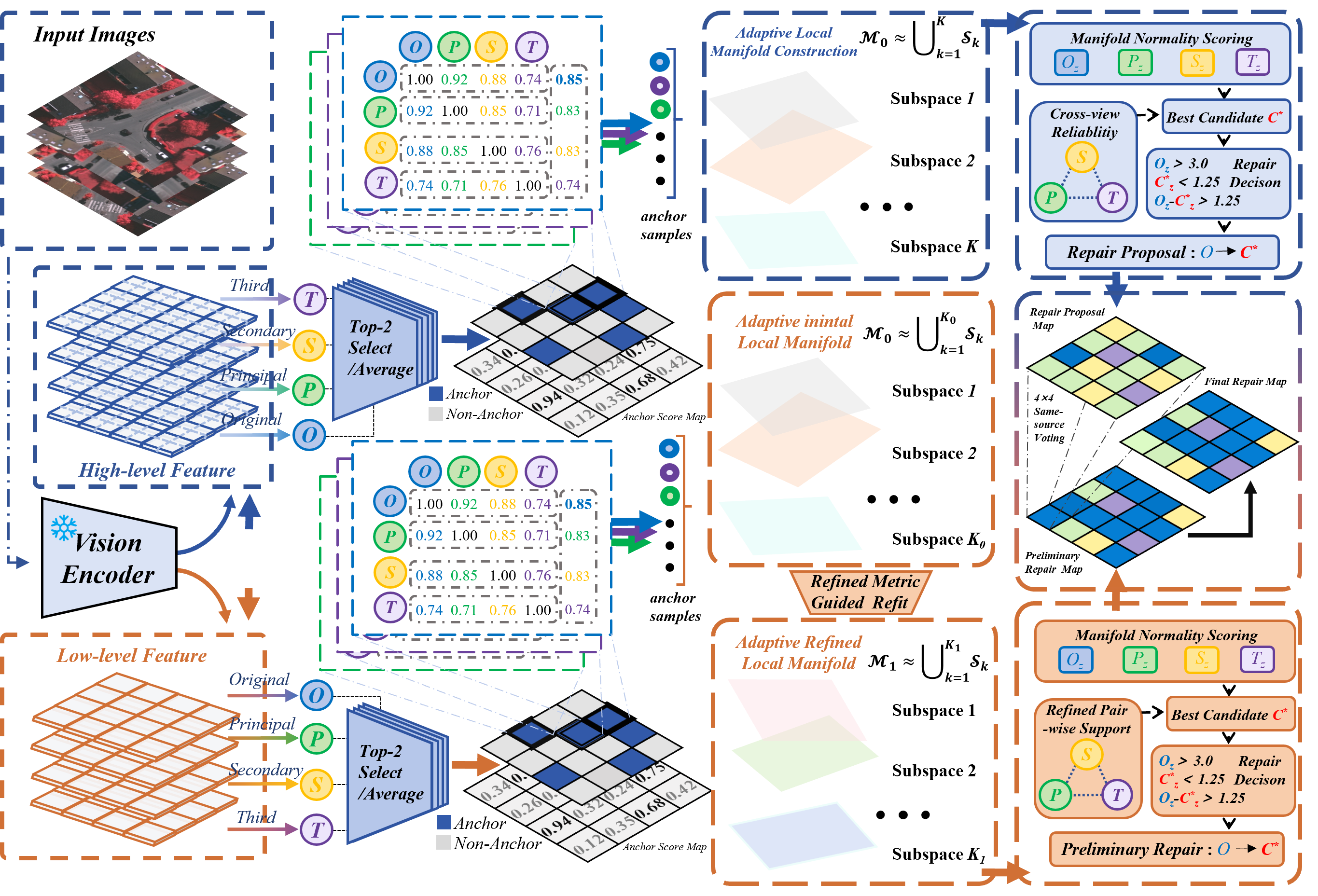}
	\caption{Overview of OAMR. Features from the three candidate views are first aligned to the original coordinate system through inverse \(\mathrm{D}_4\) transformations, and reliable cross-view observations are used to construct adaptive local subspaces. Refined cross-view metrics and candidate support then identify reliable replacement sources, enabling selective repair only of original-view features that deviate significantly from the local manifold. Finally, candidate-source consistency at a high-resolution feature level guides cross-scale extension at the target repair level.}
	\label{fig:oamr}
\end{figure*}

To address this issue, we propose Original-Anchored Adaptive Manifold Repair (OAMR) as shown in figure~\ref{fig:oamr}. OAMR first identifies reliable locations based on cross-view consistency and constructs local manifold references from actual view features. It then characterizes the multimodal structure of high-dimensional features using adaptive local subspaces and further refines cross-view reliability estimates through local tangent spaces and neighborhood relationships. Finally, complete feature replacement is performed only when a candidate view receives sufficient cross-view support and can substantially improve the manifold consistency of the original feature. Meanwhile, high-resolution feature levels provide fine-grained evidence of candidate sources to supplement local decisions at the target repair level.

\subsubsection{Reliable Anchor Initialization}
\label{subsubsec:reliable_anchor_initialization}

For the multi-scale features produced by the visual backbone, let the feature map of view \(v\) at scale \(l\) be
\begin{equation}
F_v^{(l)}\in\mathbb{R}^{C_l\times H_l\times W_l},
\label{eq:multiscale_view_features}
\end{equation}
where \(C_l\), \(H_l\), and \(W_l\) denote the number of channels, height, and width at this level, respectively. For each candidate view \(v\in\{P,S,T\}\), let \(g_v\) denote its corresponding \(\mathrm{D}_4\) transformation. We first restore the candidate features to the original coordinate system through the inverse group action:
\begin{equation}
\widetilde{F}_v^{(l)}=g_v^{-1}\!\left(F_v^{(l)}\right), \qquad v\in\{P,S,T\}.
\label{eq:inverse_feature_alignment}
\end{equation}
We also set \(\widetilde{F}_O^{(l)}=F_O^{(l)}\). The four views thus have explicit geometric correspondences at the same spatial locations. At spatial location \(x\), let \(f_v(x)\in\mathbb{R}^{C_l}\) denote the aligned feature vector of view \(v\). Cross-view consistency evaluation and local manifold modeling use its \(L_2\)-normalized form:
\begin{equation}
\widehat{f}_v(x)=\frac{f_v(x)}{\lVert f_v(x)\rVert_2+\varepsilon}.
\label{eq:normalized_view_features}
\end{equation}
This normalization is used only for reliability evaluation and manifold-distance computation. When the repair conditions are satisfied, feature replacement uses the complete, unnormalized candidate feature aligned through the inverse transformation, thereby preserving the original numerical representation of the output features.

Local manifold estimation should prioritize locations with relatively stable cross-view representations. To this end, we first compute the cosine consistency between the original view and each of the three candidate views:
\begin{equation}
c_v(x)=\widehat{f}_O(x)^{\top}\widehat{f}_v(x), \qquad v\in\{P,S,T\}.
\label{eq:original_candidate_consistency}
\end{equation}
Let \(c_{(1)}(x)\) and \(c_{(2)}(x)\) denote the two largest similarities among the three candidates. The initial anchor score is defined as
\begin{equation}
a(x)=\frac{c_{(1)}(x)+c_{(2)}(x)}{2}.
\label{eq:initial_anchor_score}
\end{equation}
We then combine an adaptive quantile threshold with a minimum consistency requirement to determine the reliable anchor set:
\begin{equation}
\left\{
\begin{aligned}
\mathcal{A}_0 &= \{x\mid a(x)\geq\tau_a\}, \\
\tau_a &= \max\!\left(Q_q(a),\tau_a^{\mathrm{min}}\right).
\end{aligned}
\right.
\label{eq:reliable_anchor_set}
\end{equation}
Here, \(Q_q(\cdot)\) denotes the quantile operator, and \(\tau_a^{\mathrm{min}}\) is the minimum consistency threshold. Combining the responses of the two most consistent candidates reduces the influence of accidentally high similarities on reliable-region identification.

At each reliable location, OAMR selects an observed feature with the highest overall consistency with the other views as the manifold reference sample, rather than averaging the features from all four views. Specifically, we define
\begin{equation}
v^{\ast}(x)=\operatorname*{arg\,max}_{v\in\mathcal{V}}\left[\frac{1}{\lvert\mathcal{V}\rvert-1}\sum_{\substack{u\in\mathcal{V}\\u\neq v}}\widehat{f}_v(x)^{\top}\widehat{f}_u(x)+\delta\,\mathbf{1}(v=O)\right].
\label{eq:reference_view_selection}
\end{equation}
Here, \(\mathbf{1}(\cdot)\) is the indicator function, and \(\delta\) introduces a small bias toward the original view, giving it higher selection priority when cross-view consistency scores are nearly tied. Each reliable location therefore contributes only one medoid drawn from the observed views to subsequent manifold estimation:
\begin{equation}
f_{\mathrm{med}}(x)=f_{v^{\ast}(x)}(x), \qquad x\in\mathcal{A}_0.
\label{eq:medoid}
\end{equation}
Selecting actual observed samples avoids synthetic representations introduced by cross-view averaging that may not belong to the actual feature distribution, while allowing stable candidate observations to contribute to the construction of local reference structures.

\subsubsection{Adaptive Local Subspace Modeling}
\label{subsubsec:adaptive_local_subspace}

High-level visual features typically contain multiple semantic regions and local structural patterns, making it difficult for a single global linear subspace to adequately characterize their distribution. OAMR constructs an adaptive mixture of local subspaces from reliable medoid samples, allowing different structural regions to be represented by different low-dimensional linear approximations.

For the \(k\)-th local subspace, let \(\mu_k\) and \(U_k\) denote its mean and orthonormal basis, respectively. Given a feature vector \(f\), its orthogonal reconstruction residual with respect to this subspace is defined as
\begin{equation}
r_k(f)=\left\lVert\left(I-U_kU_k^{\top}\right)\left(f-\mu_k\right)\right\rVert_2,
\label{eq:subspace_reconstruction_residual}
\end{equation}
where \(I\) is the identity matrix. The dimensionality of each local subspace is determined adaptively according to the cumulative explained variance of its feature spectrum, rather than being fixed in advance. This preserves the dominant feature variations while preventing excessively high-dimensional subspaces from reducing the discriminative power of anomaly residuals. To reduce the influence of a small number of anomalous anchors on manifold estimation, samples in the high-residual tail are removed during local PCA fitting, and the subspace is re-estimated. Let \(m_k\) and \(s_k\) denote the median and robust scale of the reference residuals for the \(k\)-th subspace, respectively. The standardized anomaly score of a feature relative to the local manifold is then expressed as
\begin{equation}
z(f)=\min_k\frac{r_k(f)-m_k}{s_k}.
\label{eq:standardized_manifold_anomaly}
\end{equation}
Taking the minimum ensures that each feature is evaluated against its best-matching local subspace. The robust residual scale is estimated using the residual median and the Median Absolute Deviation (MAD), reducing the influence of a small number of extreme responses on the anomaly score.

The number of local subspaces is also determined adaptively according to the feature distribution of the current image. Initially, all reliable medoid samples are used to fit a single robust local subspace. Underfitted sample clusters are then prioritized based on their upper-quantile reconstruction residuals, and candidate child clusters are generated through deterministic binary splitting. A split is accepted only when both child clusters contain sufficient samples, exhibit clear structural separation, and achieve a sufficient reduction in reconstruction error relative to the parent subspace. This process proceeds recursively until no valid split remains or a predefined complexity limit is reached. Consequently, simple scenes retain fewer local models, whereas images containing diverse structural patterns can be represented by multiple local subspaces, avoiding reliance on a fixed number of linear models.

\subsubsection{Refined Cross-View Metric-Guided Refit}
\label{subsubsec:refined_metric_refit}

Initial cosine consistency provides stable seed samples for local manifold estimation. However, pointwise directional similarity alone is insufficient to characterize anisotropic variations within the local manifold and spatial neighborhood structure. To improve the accuracy of reliable anchors and local manifold references, OAMR constructs a refined cross-view metric based on the initial mixture of local subspaces. This metric is used to reselect reliable locations and medoids from the observed views, followed by one guided refit of the local subspace mixture. This process refines the initial cosine-driven manifold estimate into a cross-view reference structure jointly constrained by local geometry and neighborhood relationships.

At location \(x\), the reference local subspace is first determined by the original view:
\begin{equation}
k_O(x)=\operatorname*{arg\,min}_k z_k\!\left(f_O(x)\right).
\label{eq:original_view_reference_subspace}
\end{equation}
Consequently, all comparisons between views at this location use the local reference structure associated with the original view, preventing any candidate view from redefining the local coordinate reference. Let \(U\) denote the orthonormal basis of the reference subspace and \(\{\lambda_j\}_{j=1}^{r}\) its retained eigenvalues. The feature difference between two aligned views \(u\) and \(v\) is defined as
\begin{equation}
\Delta_{uv}(x)=\widehat{f}_v(x)-\widehat{f}_u(x).
\label{eq:aligned_feature_difference}
\end{equation}
OAMR separately measures the variation along the local tangent space and the orthogonal component outside the local subspace, yielding the local geometric distance
\begin{equation}
d_{\mathrm{loc}}^{uv}(x)=\omega_t d_t^{uv}(x)+\omega_{\perp}d_{\perp}^{uv}(x).
\label{eq: local_geometric_distance}
\end{equation}
Here, \(d_t^{uv}\) normalizes variations along tangent-space directions according to the local eigenvalues, while \(d_{\perp}^{uv}\) measures orthogonal deviations that cannot be explained by the reference subspace. The weights \(\omega_t\) and \(\omega_{\perp}\) control the contributions of these two components and are set to \(\omega_t=0.35\) and \(\omega_{\perp}=0.65\) in this work. This formulation distinguishes anisotropic variations permitted within the local manifold from anomalous responses that deviate from the local structure.

Beyond pointwise feature differences, OAMR further characterizes the relationship pattern between each location and its local neighborhood. Let \(\pi_v(x)\) denote the local relation distribution of view \(v\) at location \(x\), obtained by normalizing the cosine relationships between the central feature and its neighboring features. The neighborhood-structure discrepancy between two views is measured using the Jensen--Shannon divergence:
\begin{equation}
d_{\mathrm{rel}}^{uv}(x)=D_{\mathrm{JS}}\!\left(\pi_u(x)\,\Vert\,\pi_v(x)\right).
\label{eq:neighborhood_relation_distance}
\end{equation}
The local geometric distance and neighborhood relation distance are standardized separately using robust statistics and then combined into the final pairwise cross-view distance:
\begin{equation}
d_{uv}(x)=Z\!\left[(1-\eta)\,Z\!\left(d_{\mathrm{loc}}^{uv}(x)\right)+\eta\,Z\!\left(d_{\mathrm{rel}}^{uv}(x)\right)\right].
\label{eq:refined_cross_view_distance}
\end{equation}
Here, \(Z(\cdot)\) denotes standardization based on the median and a robust scale estimate, and \(\eta\) controls the relative contribution of neighborhood relation information. This distance accounts for both local manifold geometry and spatial relational consistency, extending beyond pointwise cosine similarity.

Using this refined cross-view metric, OAMR reselects reliable anchors and their corresponding medoid samples, then updates the adaptive mixture of local subspaces accordingly. The updated manifold model is subsequently used to compute the final cross-view distances, providing a unified reliability criterion for candidate-support assessment and replacement-source selection.

\subsubsection{Selective Feature Replacement with Candidate Support}
\label{subsubsec:selective_feature_replacement}

Once the final local subspaces have been established, we compute the standardized residuals of the original view and the three candidate views relative to the local manifold, denoted by \(z_O(x)\), \(z_P(x)\), \(z_S(x)\), and \(z_T(x)\), respectively. A larger \(z_v(x)\) indicates a greater deviation of the corresponding feature from the current reliable local manifold. However, a low candidate residual alone does not guarantee reliability, as a candidate may produce an isolated, erroneous response that happens to lie close to a local subspace. Therefore, before admitting a candidate into the set of replacement sources, OAMR additionally requires cross-view support from other candidate views. Let \(s_v(x)\) denote the number of other candidates supporting candidate \(v\) at location \(x\). The eligible candidate set is defined as
\begin{equation}
\mathcal{E}(x)=\{v\in\{P,S,T\}\mid s_v(x)\geq\tau_s\},
\label{eq:eligible_candidate_set}
\end{equation}
where \(\tau_s\) is the required support count, set to \(\tau_s=2\) in this work. At feature levels where the refined metric is enabled, mutual support between candidates is determined by the previously defined pairwise distances and their adaptive thresholds. At higher-resolution guidance levels, support relationships are established using a more lightweight cosine-consistency measure. Candidate support determines only eligibility for replacement; it does not involve averaging multiple candidate features.

For eligible candidates, we further consider both their local manifold anomaly scores and their overall consistency across views. Let \(\bar{c}_v(x)\) denote the cross-view consistency cost of candidate \(v\), standardized using robust statistics. The candidate selection score is defined as
\begin{equation}
\psi_v(x)=(1-\rho)z_v(x)+\rho\,\bar{c}_v(x),
\label{eq:candidate_selection_score}
\end{equation}
where \(\rho\) controls the contribution of multi-view consistency to replacement-source selection. The selected candidate source is
\begin{equation}
v^{\dagger}(x)=\operatorname*{arg\,min}_{v\in\mathcal{E}(x)}\psi_v(x).
\label{eq:selected_candidate_source}
\end{equation}
If \(\mathcal{E}(x)\) is empty, the original-view feature is retained at that location. This selection process requires candidates to be both close to the reliable local manifold and consistent with other geometric observations.

Even when an eligible candidate exists, OAMR does not immediately perform replacement. A location \(x\) is identified as requiring repair only when all the following conditions are satisfied:
\begin{equation}
\left\{
\begin{aligned}
&z_O(x)>\tau_O, \\
&z_{v^{\dagger}}(x)<\tau_C, \\
&z_O(x)-z_{v^{\dagger}}(x)>\Delta_z.
\end{aligned}
\right.
\label{eq:selective_repair_conditions}
\end{equation}
Here, \(\tau_O\) identifies original-view features that deviate significantly from the reliable manifold, \(\tau_C\) ensures that the candidate lies within an acceptable manifold range, and \(\Delta_z\) requires a sufficient reduction in residual relative to the original feature. We set \(\tau_O=3.0\), \(\tau_C=1.25\), and \(\Delta_z=1.25\). The final feature update uses selective hard replacement:
\begin{equation}
f_{\mathrm{rep}}(x)=\begin{cases}
f_{v^{\dagger}}(x), & x\in\mathcal{M}_{\mathrm{rep}}, \\
f_O(x), & \text{otherwise},
\end{cases}
\label{eq:selective_hard_replacement}
\end{equation}
where \(\mathcal{M}_{\mathrm{rep}}\) denotes the initial set of repair locations satisfying the above conditions. The replacement feature \(f_{v^{\dagger}}(x)\) is the complete, unnormalized candidate feature aligned through the inverse transformation, rather than the normalized vector used for scoring. Thus, OAMR neither computes a weighted average of candidate features nor continuously interpolates between the original and candidate representations. Each repaired location retains a complete high-dimensional feature produced by an actual view.

\subsubsection{Cross-Scale Guidance}
\label{subsubsec:cross_scale_guidance}

Visual features at the target repair level have relatively low spatial resolution, with each feature location corresponding to a comparatively large image region. Consequently, they may provide only a coarse characterization of local anomalies and fine-grained spatial structures. OAMR therefore introduces cross-scale guidance, using a high-resolution feature level to provide finer-grained evidence of candidate sources and supplement local decisions at the target repair level.

Let \(l_r\) denote the target level where feature repair is performed, and let \(l_g\) denote the corresponding guidance level with higher spatial resolution. Both levels independently determine candidate sources, but the guidance level only provides cross-scale evidence; its features are not directly written into the final output. Because the two levels have explicit spatial correspondences, each location \(x\) at the target repair level corresponds to a local region \(B(x)\) at the high-resolution guidance level. Let \(S_g(y)\) denote the candidate-source mapping obtained at the guidance level, where \(y\in B(x)\) represents a high-resolution location within the corresponding region. For candidate view \(v\in\{P,S,T\}\), its source-support count within \(B(x)\) is defined as
\begin{equation}
N_v(x)=\sum_{y\in B(x)}\mathbf{1}\!\left[S_g(y)=v\right].
\label{eq:cross_scale_support_count}
\end{equation}
Here, \(\mathbf{1}(\cdot)\) is the indicator function. We then select the candidate view receiving the most fine-grained support:
\begin{equation}
v_{\mathrm{cs}}(x)=\operatorname*{arg\,max}_{v\in\{P,S,T\}}N_v(x).
\label{eq:cross_scale_candidate_source}
\end{equation}
This view is the most strongly supported candidate within the corresponding high-resolution region. When its support count reaches a predefined threshold, location \(x\) is considered to have stable cross-scale candidate evidence. This process aggregates consistent candidate-source decisions from multiple high-resolution locations into the corresponding location at the target repair level, reducing the influence of an individual coarse-resolution feature response on the repair decision.

High-resolution evidence is used only to supplement locations that have not yet been selected for repair by the local manifold criterion at the target level. It does not reverse replacements already accepted at that level. Let \(\mathcal{M}_{\mathrm{rep}}\) denote the initial repair set and \(\mathcal{M}_{\mathrm{cs}}\) the set of additional locations supported by cross-scale evidence. The final repair set is
\begin{equation}
\mathcal{M}_{\mathrm{final}}=\mathcal{M}_{\mathrm{rep}}\cup\mathcal{M}_{\mathrm{cs}}.
\label{eq:final_repair_set}
\end{equation}
For locations \(x\in\mathcal{M}_{\mathrm{rep}}\), the candidate source determined by the local manifold criterion at the target repair level is retained. For locations \(x\in\mathcal{M}_{\mathrm{cs}}\) added solely through cross-scale evidence, replacement uses the complete candidate feature corresponding to \(v_{\mathrm{cs}}(x)\) at the target repair level, aligned to the original coordinate system. Thus, high-resolution guidance extends the coverage of reliable repair through source consistency within fine-grained regions while preserving existing repair decisions.

Finally, only the repaired features at the target repair level are written into the output feature state. The high-resolution guidance level and all other feature levels where repair is disabled retain their original-view representations. Positional encodings and other deterministic spatial information are likewise inherited from the original view and are excluded from local manifold estimation and cross-view replacement, ensuring that the repaired features maintain a spatial reference consistent with the original image.

Through this cross-scale guidance mechanism, OAMR converts fine-grained candidate consistency in high-resolution features into additional reliability evidence for the target repair level. This allows locations that lack sufficient evidence to trigger repair at a single scale to be corrected when stable cross-view support exists within their corresponding local regions. Combined with the preceding original-view-anchored local manifold modeling and candidate-support constraints, OAMR selectively uses complementary information from multiple views without indiscriminately fusing entire feature maps.

\subsection{Pixel Decoder Test-Time Adaptation}
\label{subsec:pd_tta}

\begin{figure}
	\centering
	\includegraphics[width=.45\textwidth]{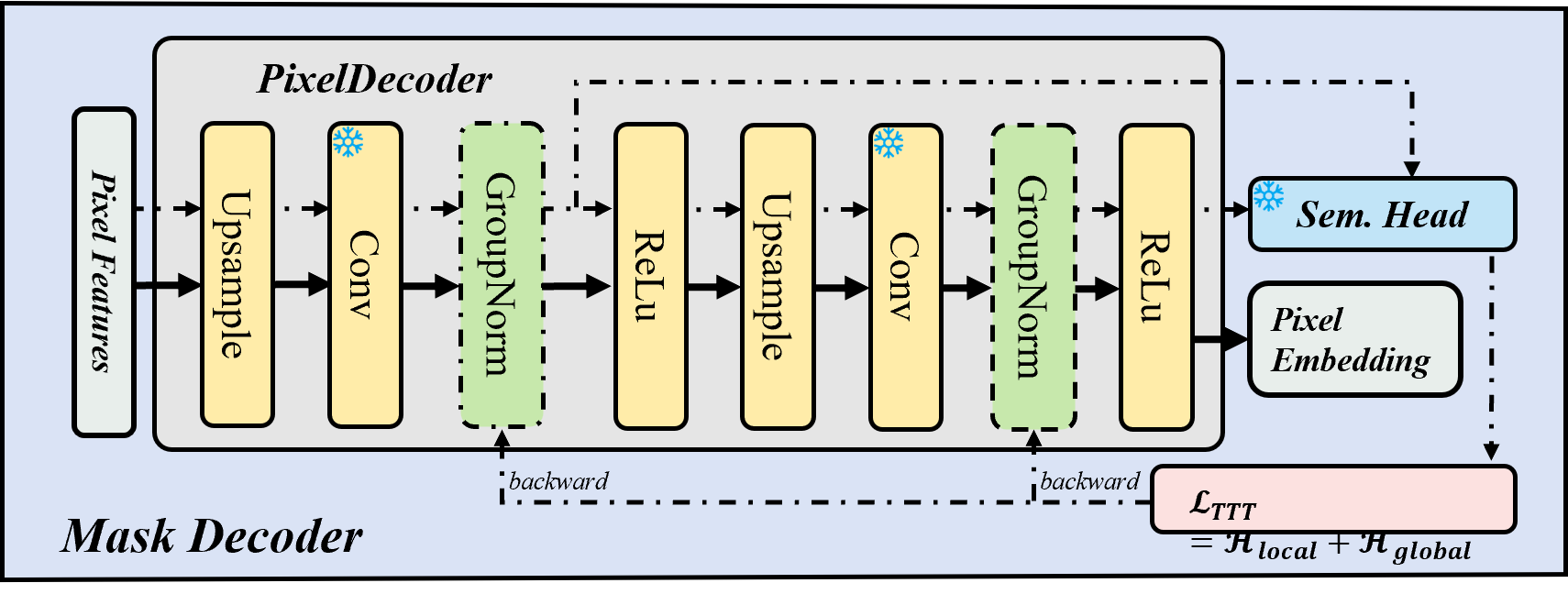}
	\caption{The PD-TTA in PixelDecoder of SAM3}
	\label{fig:pdtta}
\end{figure}

The preceding multi-view selection and manifold repair enhance the adaptability of SAM3 to geometric transformations in remote sensing imagery at the input-observation and visual-representation levels, respectively. The repaired visual features continue through the original cross-modal encoding, MaskDecoder, and PixelDecoder pathways of SAM3 to produce pixel-level predictions. However, the affine parameters of the GroupNorm layers in the PixelDecoder are pretrained on source-domain data, and their normalization behavior may not adequately accommodate variations in appearance and scene distributions across remote sensing images. We therefore further propose Pixel Decoder Test-Time Adaptation (PD-TTA) as shown in figure~\ref{fig:pdtta} for SAM3.

PD-TTA introduces no separate adaptation branch and preserves the original input and output formats of SAM3. Instead, it directly designates the affine parameters of the GroupNorm layers that participate in the PixelDecoder forward pass as the parameters to be adapted at test time. An unsupervised loss is then constructed from the responses produced by the original semantic prediction pathway of SAM3.

\subsubsection{Integration with the SAM3 PixelDecoder}
\label{subsubsec:pd_tta_integration}

For the current test image, we follow the original text-prompting mechanism of SAM3 and sequentially use the dataset category names as query words. Text encoding and cross-modal interaction produce category-specific visual--semantic features, which are then passed to the PixelDecoder through the original decoding pathway. In the current model configuration, two GroupNorm layers in the PixelDecoder actively participate in the forward computation. During test-time adaptation, only the affine scale parameters \(\gamma\) and bias parameters \(\beta\) of these GroupNorm layers receive gradients. The convolutional layers, text-related modules, semantic head, and all other learnable parameters remain frozen.

PD-TTA therefore adjusts the normalization behavior within the existing PixelDecoder without replacing any network layers in SAM3, allowing the adapted features to pass seamlessly through its original subsequent decoding and prediction stages. To reduce the computational cost of backpropagation, only the feature copies used for adaptation are resized while preserving their aspect ratio. The visual states and spatial dimensions used for the actual inference pass remain unchanged.

\subsubsection{Semantic Feedback and Local--Global Entropy Objective}
\label{subsubsec:semantic_feedback_entropy}

To directly constrain PixelDecoder parameter updates using the semantic predictions of SAM3, we decode features from different active stages of the PixelDecoder using the frozen semantic head. Semantic responses at different spatial scales are aligned to the resolution of the final active stage through bilinear interpolation and then fused with equal weights. For category \(c\), the fused semantic response is expressed as
\begin{equation}
\left\{
\begin{aligned}
\bar{z}_c(u;\theta)&=\frac{1}{S}\sum_{s=1}^{S}\left[\mathcal{I}_s\!\left(H\!\left(D_{\theta}^{(s)}(F_c)\right)\right)\right]_u, \\
\theta&=\{\gamma_l,\beta_l\}_{l\in\mathcal{G}}.
\end{aligned}
\right.
\label{eq:fused_semantic_response}
\end{equation}
Here, \(c\) denotes the category index, and \(F_c\) denotes the PixelDecoder input features obtained using the name of category \(c\) through the original text-prompting and cross-modal encoding pathways of SAM3. The quantity \(S\) is the number of PixelDecoder feature levels used for semantic-response fusion, and \(s\) indexes these levels. The mapping \(D_{\theta}^{(s)}(\cdot)\) denotes the forward computation producing features at the \(s\)-th PixelDecoder level under parameters \(\theta\). The function \(H(\cdot)\) is the frozen semantic head, while \(\mathcal{I}_s(\cdot)\) aligns the semantic response from the \(s\)-th feature level to the common spatial resolution through bilinear interpolation. The variable \(u\) denotes a spatial location in the fused feature map, and \(\bar{z}_c(u;\theta)\) is the fused semantic logit for category \(c\) at that location. The set \(\theta\) contains the parameters adapted at test time, consisting of the affine scale parameters \(\gamma_l\) and bias parameters \(\beta_l\) of the active GroupNorm layers indexed by \(\mathcal{G}\). This formulation specifies how PD-TTA is integrated into SAM3: the adaptable parameters reside within the PixelDecoder, while the semantic feedback driving adaptation comes from the original semantic head.

After obtaining the fused logits, we apply softmax along the category dimension to obtain pixel-level category probabilities. Image-level category probabilities are obtained by first spatially averaging the fused logits and then applying softmax along the category dimension. To jointly constrain local pixel predictions and image-level category discrimination, and inspired by the joint local--global entropy optimization strategy used in the MLMP test-time adaptation method \cite{noori_test-time_2025}, PD-TTA adopts the following local--global entropy objective:
\begin{equation}
\mathcal{L}_{\mathrm{TTA}}=-\frac{1}{\lvert\Omega\rvert}\sum_{u\in\Omega}\sum_{c=1}^{C}p_{c,u}\log p_{c,u}-\alpha_{\mathrm{cls}}\sum_{c=1}^{C}q_c\log q_c.
\label{eq:local_global_entropy_loss}
\end{equation}
Here, \(\mathcal{L}_{\mathrm{TTA}}\) is the unsupervised test-time adaptation loss of PD-TTA. The quantity \(C\) denotes the total number of dataset categories, with \(c\in\{1,\ldots,C\}\) indexing the categories. The set \(\Omega\) contains all spatial locations in the fused feature map, \(\lvert\Omega\rvert\) is the number of these locations, and \(u\in\Omega\) denotes an individual location. The probability \(p_{c,u}\) is the pixel-level prediction that location \(u\) belongs to category \(c\), obtained by applying softmax to the fused logits along the category dimension. Thus, the first term on the right-hand side is the mean local entropy over all spatial locations.

The probability \(q_c\) is the image-level category probability obtained by spatially averaging the fused logits over \(\Omega\) and then applying softmax along the category dimension. Accordingly, the second term is the weighted global category entropy. The coefficient \(\alpha_{\mathrm{cls}}\) controls the contribution of this global entropy term and balances the local pixel-level and image-level category constraints. We set \(\alpha_{\mathrm{cls}}=1.0\) in this work.

\subsubsection{Test-Time Adaptation and Inference}
\label{subsubsec:tta_inference}

At test time, the entropy loss is computed from the semantic feedback of SAM3 described above. Backpropagation computes gradients only for the GroupNorm parameters \(\gamma\) and \(\beta\) within the PixelDecoder, which are updated using the Adam optimizer. Throughout optimization, the text encoder, cross-modal interaction modules, convolutional layers, semantic head, and all other parameters of SAM3 remain frozen. After the predefined number of test-time updates, the adapted GroupNorm parameters are used directly in the existing PixelDecoder, and the actual inference pass for the current image follows the original cross-modal encoding, MaskDecoder, and semantic prediction pathways of SAM3. PD-TTA therefore confines adaptation to normalization within the PixelDecoder while preserving the original prompting mechanism, semantic decoding architecture, and inference pipeline.

The key experimental settings are \(\eta=1\times10^{-3}, \qquad T=1, \qquad S_{\max}=72, \qquad \alpha_{\mathrm{cls}}=1.0\). Here, \(\eta\) is the Adam learning rate used to update the GroupNorm affine parameters at test time. The quantity \(T\) is the number of gradient updates performed for the current test image. The parameter \(S_{\max}\) specifies the maximum spatial side length allowed when resizing the adaptation feature copies while preserving their aspect ratio to reduce backpropagation costs. The coefficient \(\alpha_{\mathrm{cls}}\) is the weight of the global category entropy term in the local--global entropy objective. These settings control only the strength and computational cost of PixelDecoder normalization adaptation for the current image; they do not alter the input dimensions used for actual inference or the original network architecture of SAM3.

\section{Experiments}
\label{sec:experiments}

We evaluate the proposed method on eight widely used remote sensing semantic segmentation datasets: OpenEarthMap \cite{xia_openearthmap_2023}, LoveDA \cite{wang_loveda_2021}, iSAID \cite{zamir_isaid_2019}, UAVid \cite{lyu_uavid_2020}, UDD5 \cite{chen_large-scale_2018}, VDD \cite{cai_vdd_2025}, Potsdam and Vaihingen \cite{rottensteiner_isprs_2012}. Mean Intersection-over-Union (mIoU) is used as the main evaluation metric across all datasets.

\subsection{Datasets}
\label{subsec:datasets}
The introduction of the employed eight remote sensing datasets are shown as follows: 

(1) OpenEarthMap consists of high-resolution remote sensing images acquired from aerial and satellite platforms, covering 44 countries and regions worldwide. The imagery has a spatial resolution of approximately 0.25--0.50~m, primarily contains three visible RGB bands, and provides annotations for eight land-cover classes.

(2) LoveDA contains high-resolution remote sensing images obtained through Google Earth, with a spatial resolution of approximately 0.30~m and three visible RGB bands. It covers both urban and rural scenes and includes seven land-cover classes.

(3) iSAID is built upon the high-resolution aerial imagery of DOTA-v1.0. Its images originate from multiple sources, including Google Earth, GF-2, and JL-1, with spatial resolutions varying by source. The dataset primarily contains RGB imagery, together with some panchromatic imagery, and provides annotations for 15 typical object categories.

(4) UAVid consists of high-resolution urban video imagery captured from oblique viewpoints using unmanned aerial vehicles (UAVs), primarily in RGB format. Because the imagery is acquired through low-altitude oblique imaging, its ground spatial resolution varies with the imaging distance and viewing angle. The dataset includes eight categories of urban land cover and objects.

(5) UDD5 contains three-channel RGB imagery acquired by UAVs across multiple urban and campus scenes at flight altitudes of approximately 60--100~m. It comprises five classes: vegetation, buildings, roads, vehicles, and others.

(6) VDD contains high-resolution RGB imagery captured using a DJI Mavic Air II UAV across urban, industrial, rural, and natural scenes. Its ground spatial resolution varies with viewing angle and flight conditions. The dataset includes seven land-cover classes and covers diverse viewing angles and illumination conditions.

(7) Potsdam and (8) Vaihingen are both part of the ISPRS 2D Semantic Labeling Benchmark and consist of very-high-resolution orthophotos acquired from aerial platforms. Their spatial resolutions are approximately 0.05~m and 0.099~m, respectively. Potsdam provides four-band imagery comprising RGB and near-infrared (NIR) bands, whereas Vaihingen primarily contains NIR, red, and green (NIR--R--G) bands. Both datasets include six urban land-cover classes.

Together, these datasets span satellite, aerial, and UAV imaging platforms and exhibit substantial differences in spatial resolution, object scale, and viewpoint. They therefore enable a comprehensive evaluation of OVSS performance across diverse remote sensing scenarios.

\subsection{Implementation Details}
\label{subsec:implementation_details}

We use the officially released SAM3 model with its Perception Encoder-Large+ (PE-L+) visual backbone and follow the text-prompting and prediction-fusion settings of SegEarth-OV3 \cite{li_segearth-ov3_2025} by default. GFMR requires no dataset-level training on the target datasets. Except for the GroupNorm affine parameters updated online by PD-TTA during prediction for each individual image, the visual encoder, text encoder, PixelDecoder convolutional layers, instance head, semantic head, and all other module parameters of SAM3 remain frozen. The original mask-generation and result-fusion procedures are preserved.

MH-D4VS and OAMR use the same inference configuration across all datasets, without parameter adjustments for individual test sets. MH-D4VS performs candidate geometric view selection before visual encoding, while OAMR operates on the aligned multi-view visual features and preserves the spatial reference of the original view and the subsequent prediction pipeline. None of them requires gradient computation or model parameter updates; together, they form the training-free visual adaptation core of GFMR.

The PD-TTA settings are fixed across all experiments as \(\eta=1\times10^{-3}\), \(T=1\), \(S_{\max}=72\), and \(\alpha_{\mathrm{cls}}=1.0\). A new Adam optimizer is initialized for each image, and the SAM3's default GroupNorm parameters are restored after the actual prediction pass. Consequently, neither adapted parameters nor optimizer momentum is carried over between images.

\begin{figure*}
	\centering
	\includegraphics[width=1.0\textwidth]{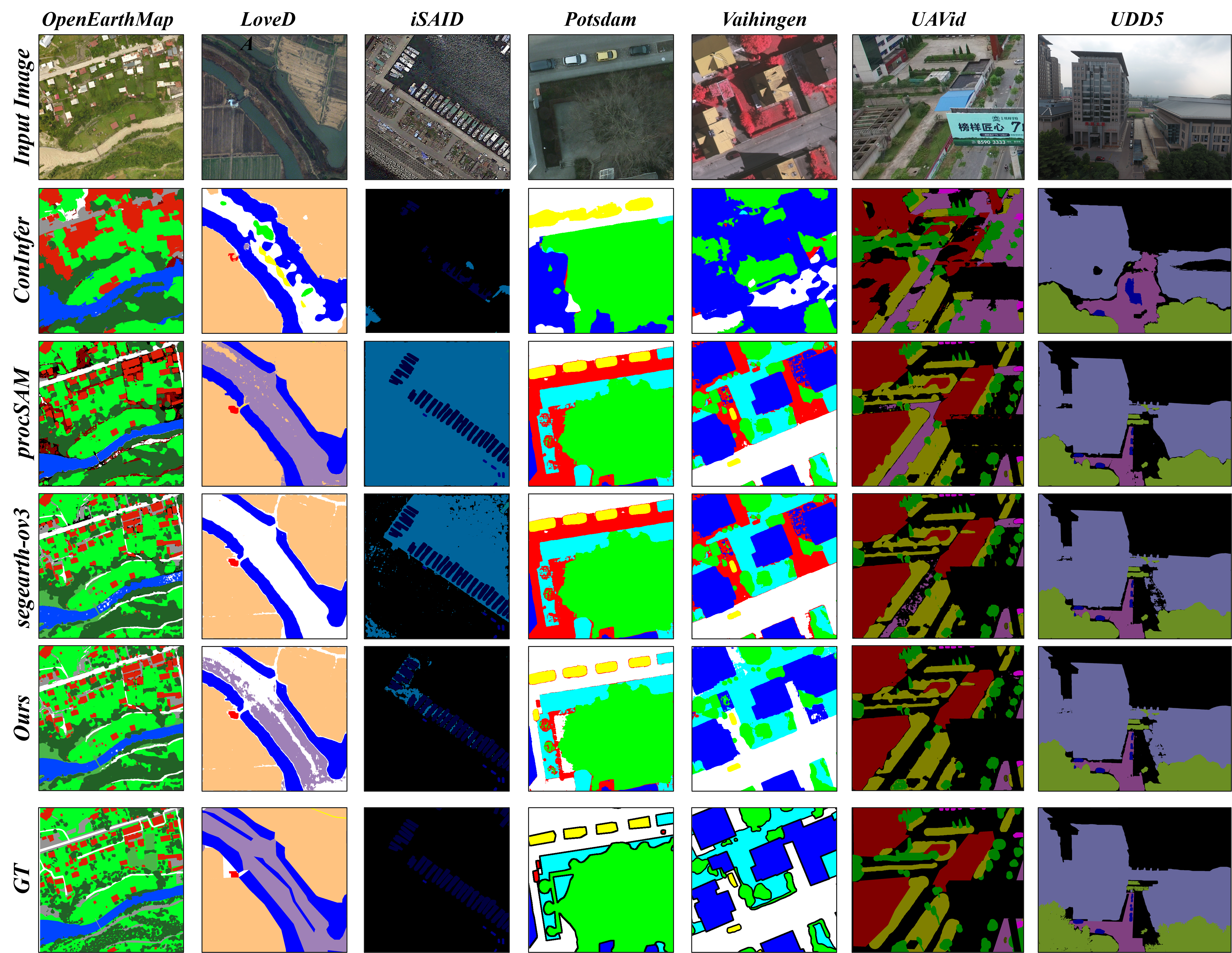}
	\caption{Qualitative comparison of the proposed method with existing open-vocabulary semantic segmentation methods.}
	\label{fig:qualitative_comparison}
\end{figure*}

\begin{table*}[t]
\centering
\caption{Performance comparison of open-vocabulary semantic segmentation methods on eight remote sensing benchmark datasets (mIoU, \%). The best results are highlighted in bold.}
\label{tab:main_comparison}

\fontsize{7}{7.2}\selectfont
\setlength{\tabcolsep}{2pt}
\renewcommand{\arraystretch}{1.15}

\begin{tabular*}{\textwidth}{
    @{\extracolsep{\fill}}
    >{\fontsize{7}{7.2}\selectfont}l
    *{9}{>{\fontsize{7}{7.2}\selectfont}c}
    @{}
}
\toprule
Method & OpenEarthMap & LoveDA & iSAID & Potsdam & Vaihingen & UAVid & UDD5 & VDD & Avg. \\
\midrule
CLIP \cite{radford_learning_2021} & 12.0 & 12.4 & 7.5 & 15.6 & 10.8 & 10.9 & 9.5 & 14.2 & 11.6 \\

MaskCLIP \cite{zhou_extract_2022} & 25.1 & 27.8 & 14.5 & 33.9 & 29.9 & 28.6 & 32.4 & 32.9 & 28.1 \\

SCLIP \cite{wang_sclip_2025} & 29.3 & 30.4 & 16.1 & 39.6 & 35.9 & 31.4 & 38.7 & 37.9 & 32.4 \\

GEM \cite{bousselham_grounding_2024} & 33.9 & 31.6 & 17.7 & 39.1 & 36.4 & 33.4 & 41.2 & 39.5 & 34.1 \\

ClearCLIP \cite{lan_clearclip_2025} & 31.0 & 32.4 & 18.2 & 42.0 & 36.2 & 36.2 & 41.8 & 39.3 & 34.6 \\

ProxyCLIP \cite{lan_proxyclip_2024} & 38.9 & 34.3 & 21.8 & 49.0 & 47.5 & 35.8 & 40.8 & 47.8 & 39.5 \\

ResCLIP \cite{yang_resclip_2024} & 39.8 & 35.5 & 21.2 & 47.4 & 40.3 & 42.5 & 50.4 & 42.5 & 39.9 \\

SegEarth-OV \cite{li_segearth-ov_2025} & 40.3 & 36.9 & 21.7 & 47.1 & 40.0 & 42.5 & 50.6 & 45.3 & 40.6 \\

CorrCLIP \cite{zhang_corrclip_2025} & 32.9 & 36.9 & 25.5 & 51.9 & 47.0 & 38.3 & 46.1 & 47.3 & 40.7 \\

ReAttnCLIP \cite{niu_reattnclip_2026} & 41.1 & 37.0 & 23.2 & 48.7 & 29.9 & 44.0 & 53.7 & 49.7 & 40.9 \\

ConInfer \cite{chen_coninfer_2026} & 42.0 & 39.3 & 20.1 & 50.0 & 31.4 & 46.4 & 46.9 & 50.3 & 40.8 \\

ProC-SAM3 \cite{song_prompt-calibrated_2026} & 41.0 & 41.6 & 20.7 & 57.1 & 60.4 & 54.3 & 71.6 & 64.2 & 51.4 \\

SegEarth-OV3 \cite{li_segearth-ov3_2025} & 42.9 & 47.4 & 27.6 & 57.8 & 60.8 & 54.7 & 71.7 & 64.5 & 53.4 \\

\midrule
\textbf{Ours} & \textbf{49.5} & \textbf{48.3} & \textbf{29.9} & \textbf{58.4} & \textbf{64.3} & \textbf{55.3} & \textbf{72.7} & \textbf{66.2} & \textbf{55.6} \\
\bottomrule
\end{tabular*}
\end{table*}

\subsection{Comparison with State-of-the-Art Methods}
\label{sec:comparision}

Figure~\ref{fig:qualitative_comparison} presents a qualitative comparison between GFMR and existing OVSS methods. Table~\ref{tab:main_comparison} reports the corresponding quantitative results on eight remote sensing benchmark datasets. GFMR achieves an average mIoU of 55.6\%, outperforming all compared methods. These results demonstrate the complementary benefits of geometric view selection, anchored manifold repair, and sample-specific decoder adaptation while preserving the core architecture and original prediction pipeline of SAM3 and keeping all model parameters frozen except the GroupNorm affine parameters updated by PD-TTA.

Across individual datasets, GFMR maintains consistent advantages under different remote sensing imaging conditions. On OpenEarthMap, the mIoU increases from 42.9\% with SegEarth-OV3 to 49.5\%. On Vaihingen, which contains high-resolution aerial imagery, it improves from 60.8\% to 64.3\%. GFMR also achieves 72.7\% and 66.2\% on UDD5 and VDD, respectively. Improvements of varying magnitudes are also observed on LoveDA, iSAID, Potsdam, and UAVid. These results indicate that the benefits extend across satellite, aerial, and UAV platforms rather than being restricted to a single data distribution.

GFMR consists of two complementary adaptation mechanisms applied at test time. MH-D4VS and OAMR adjust only geometric views and visual feature states, without gradient computation or model parameter updates. PD-TTA requires neither target-dataset annotations nor offline training and performs self-supervised updates only to the affine parameters of the GroupNorm layers in the PixelDecoder when predicting each test image. GFMR thus retains the advantages of training-free visual adaptation while introducing lightweight, image-specific online adaptation to further improve robustness to sample-level distribution shifts.

To evaluate the compatibility of the training-free visual adaptation core with different SAM3-based inference frameworks, we integrate MH-D4VS and OAMR into ProC-SAM3 \cite{song_prompt-calibrated_2026}, preserving its original text prompts, prediction heads, and subsequent decoding pipeline. PD-TTA is excluded from this transfer experiment to isolate the effects of transferring the two modules from those of additional test-time parameter updates. Table~\ref{tab:proc_sam3_transfer} reports the corresponding quantitative results.

\begin{table*}[t]
\centering
\caption{Evaluation of the proposed training-free visual adaptation modules on ProC-SAM3, another SAM~3-based framework (mIoU, \%).}
\label{tab:proc_sam3_transfer}

\fontsize{7}{7.2}\selectfont
\setlength{\tabcolsep}{2pt}
\renewcommand{\arraystretch}{1.15}

\begin{tabular*}{\textwidth}{
    @{\extracolsep{\fill}}
    >{\fontsize{7}{7.2}\selectfont}l
    *{9}{>{\fontsize{7}{7.2}\selectfont}c}
    @{}
}
\toprule
Method & OpenEarthMap & LoveDA & iSAID & Potsdam & Vaihingen & UAVid & UDD5 & VDD & Avg. \\
\midrule
ProC-SAM3
& 41.0 & 41.6 & 20.7 & 57.1 & 60.4 & 54.3 & 71.6 & 64.1 & 51.4 \\

\textbf{ProC-SAM3 + MH-D4VS + OAMR}
& \textbf{47.7} & \textbf{42.0} & \textbf{25.8}
& \textbf{57.3} & \textbf{60.6} & \textbf{54.6}
& \textbf{72.4} & \textbf{65.4} & \textbf{53.2} \\
\bottomrule
\end{tabular*}

\end{table*}

Integrating MH-D4VS and OAMR increases the average mIoU of ProC-SAM3 from 51.4\% to 53.2\%, with positive gains on all eight datasets. OpenEarthMap and iSAID improve by 6.7 and 5.1 percentage points, respectively, while Potsdam, Vaihingen, UAVid, UDD5, and VDD also show improvements. These results indicate that visual representation enhancement through geometric view selection and anchored manifold repair does not depend on a specific implementation of text prompting or result fusion and can be transferred to other SAM3-based OVSS frameworks.

\begin{figure}
    \centering
    \includegraphics[width=.48\textwidth]{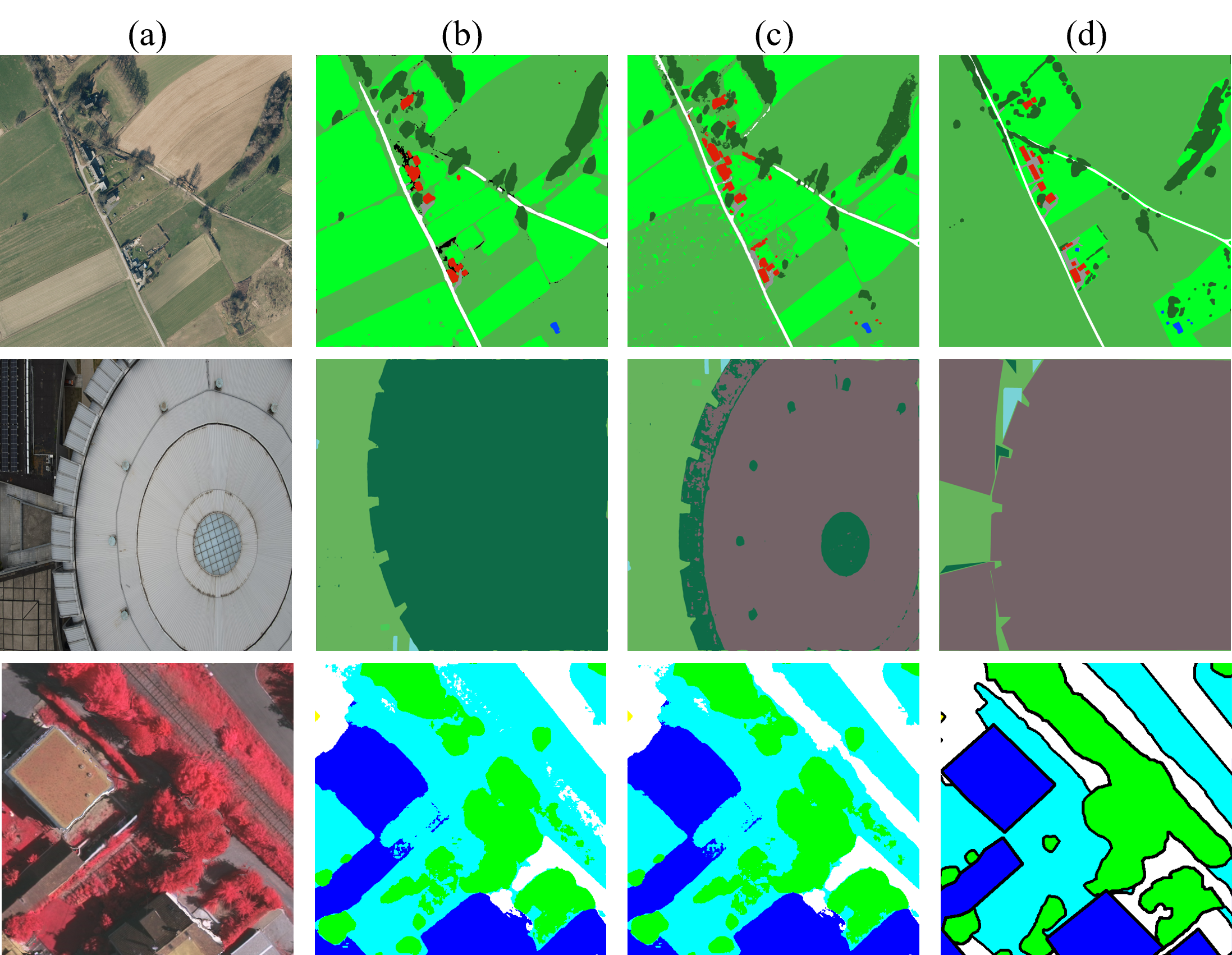}
    \caption{Qualitative results of the proposed method within the ProC-SAM3 framework. (a) Input images; (b) predictions from the original ProC-SAM3; (c) predictions after integrating the proposed MH-D4VS and OAMR modules; (d) ground truth.}
    \label{fig:proc_sam3_qualitative}
\end{figure}

Figure~\ref{fig:proc_sam3_qualitative} further presents a qualitative comparison. Compared with the original ProC-SAM3, integrating the proposed modules produces more complete object coverage and fewer local misclassifications in some regions. In particular, the second row shows numerous misclassifications in the original ProC-SAM3 predictions relative to the ground truth, which are substantially reduced after incorporating MH-D4VS and OAMR. Together with the quantitative results, these observations indicate that MH-D4VS and OAMR, as training-free modules operating within the visual representation pathway, are compatible with different SAM3-based inference frameworks.

\subsection{Ablation Studies}
\label{subsec:ablation_studies}

\subsubsection{Overall Module Analysis}
\label{subsubsec:overall_module_analysis}

We use the complete inference pipeline of SegEarth-OV3 as the baseline and introduce MH-D4VS, OAMR, and PD-TTA in different configurations to analyze their contributions to overall performance. Random-3 replaces the structure-aware triplet selection of MH-D4VS, while Mean Fusion replaces the selective feature repair of OAMR. Random-3 randomly selects three views without replacement from the seven non-identity candidate transformations, leaving the remaining visual encoding and feature alignment procedures unchanged. Mean Fusion applies the inverse transformations to align the three candidate feature maps and then averages them with the original-view features using equal weights. Starting from the SAM3's default GroupNorm parameters, PD-TTA performs one online, sample-specific update based on the local--global semantic entropy of the current image and uses the adapted state for inference. All other model parameters remain frozen.

\begin{table*}[t]
\centering
\caption{Overall ablation results for MH-D4VS, OAMR, and PD-TTA (mIoU, \%). A dash (--) indicates that the original SegEarth-OV3 baseline configuration is retained without introducing the corresponding module or replacement strategy. A check mark (\(\checkmark\)) indicates the proposed module. For MH-D4VS, a cross (\(\times\)) denotes Random-3 view selection, while a circle (\(\circ\)) denotes the use of all candidate views. For OAMR, a cross (\(\times\)) denotes Mean Fusion. PD-TTA denotes Pixel Decoder Test-Time Adaptation.}
\label{tab:overall_ablation}

\fontsize{7}{7.2}\selectfont
\setlength{\tabcolsep}{2pt}
\renewcommand{\arraystretch}{1.15}

\begin{tabular*}{\textwidth}{
    @{\extracolsep{\fill}}
    >{\fontsize{7}{7.2}\selectfont}l
    *{6}{>{\fontsize{7}{7.2}\selectfont}c}
    @{}
}
\toprule
Ablation Setting & MH-D4VS & OAMR & PD-TTA & OpenEarthMap & Vaihingen & VDD \\
\midrule
\textbf{\textit{A.}} (SegEarth-OV3 baseline)
& -- & -- & -- & 42.90 & 60.80 & 64.50 \\

\textbf{\textit{B.}}
& \(\checkmark\) & \(\times\) & -- & 44.16 & 60.77 & 53.05 \\

\textbf{\textit{C.}}
& \(\times\) & \(\checkmark\) & -- & 46.27 & 62.74 & 64.79 \\
\textbf{\textit{D.}}
& \(\circ\) & \(\checkmark\) & -- & 47.21 & 63.17 & 65.06 \\
\textbf{\textit{E.}}
& \(\checkmark\) & \(\checkmark\) & --
& 48.78 & 63.79 & 65.77 \\

\textbf{\textit{F.}}
& \(\checkmark\) & \(\checkmark\) & \(\checkmark\)
& \textbf{49.45} & \textbf{64.30} & \textbf{66.21} \\
\bottomrule
\end{tabular*}

\end{table*}

Table~\ref{tab:overall_ablation} shows that the complete GFMR framework (setting F) achieves the best results on OpenEarthMap, Vaihingen, and VDD. Compared with the baseline (setting A), it improves mIoU by 6.55, 3.50, and 1.71 percentage points, respectively. These results demonstrate that the three modules provide complementary benefits through geometric observation construction, feature reliability repair, and online calibration of sample-level distributions.

With MH-D4VS retained, replacing OAMR with Mean Fusion (setting B) substantially degrades performance, particularly on VDD, indicating that indiscriminate averaging can propagate locally inconsistent responses. With OAMR retained, replacing MH-D4VS with Random-3 (setting C) also underperforms structure-aware selection, highlighting the importance of representative and complementary candidate views for effective subsequent repair. Using all candidate views (setting D) yields modest gains over Random-3, but still underperforms the proposed structure-aware selection, suggesting that simply increasing the number of views does not guarantee more effective complementary information. Adding PD-TTA to the training-free core (setting F) further improves performance on all three datasets, demonstrating that a single image-specific online adaptation step based on entropy minimization provides consistent additional benefits.

\begin{figure*}
	\centering
	\includegraphics[width=1.0\textwidth]{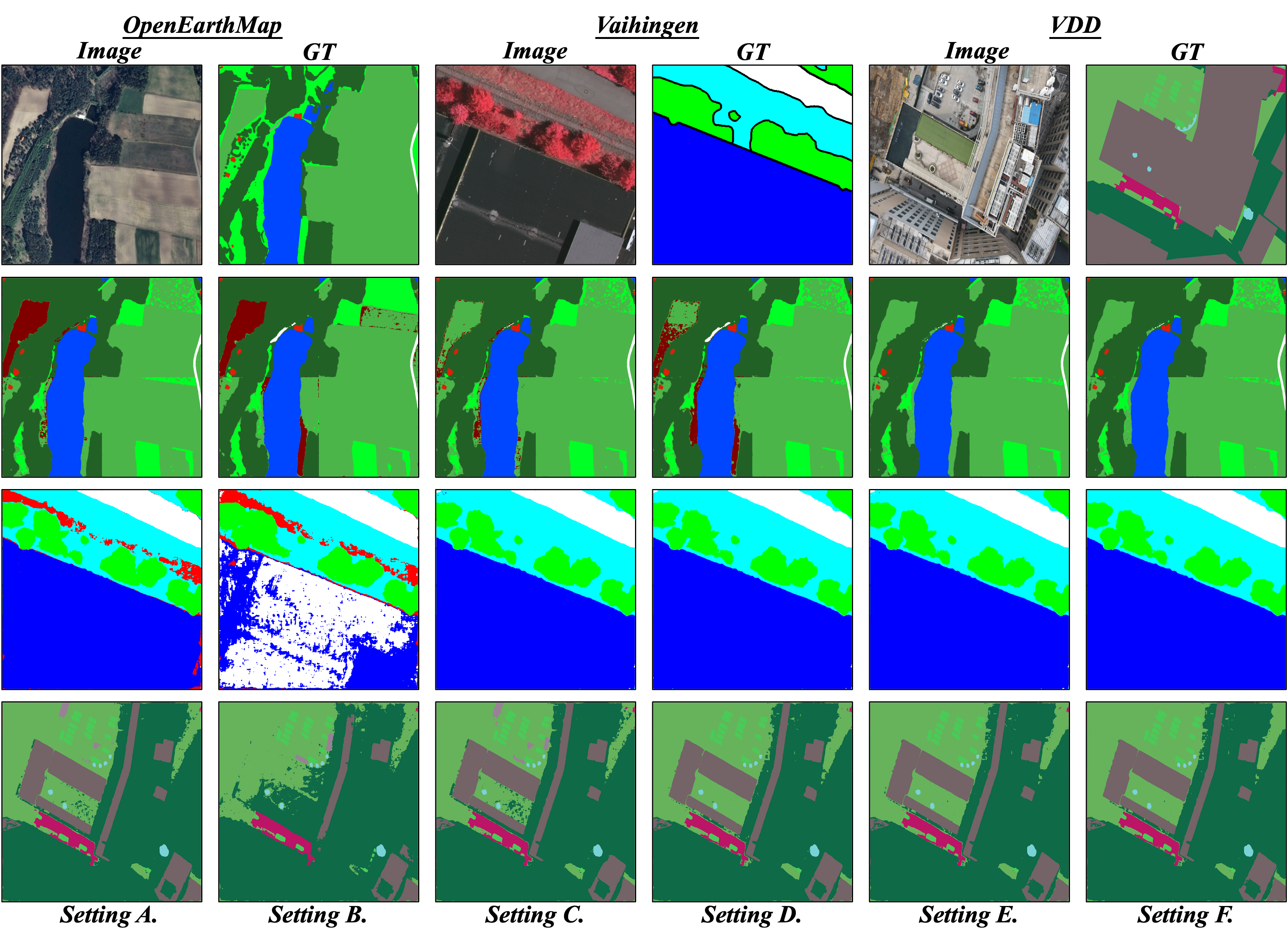}
	\caption{Qualitative comparison of different ablation configurations. From left to right: representative samples from the OpenEarthMap, Vaihingen, and VDD datasets.}
	\label{fig:abs}
\end{figure*}

\subsubsection{Ablation of MH-D4VS}
\label{subsubsec:mh_d4vs_ablation}

\begin{table*}[t]
\centering
\caption{Ablation study of different components in MH-D4VS (mIoU, \%).}
\label{tab:mh_d4vs_ablation}

\fontsize{7}{7.2}\selectfont
\setlength{\tabcolsep}{2pt}
\renewcommand{\arraystretch}{1.15}

\begin{tabular*}{\textwidth}{
    @{\extracolsep{\fill}}
    *{7}{>{\fontsize{7}{7.2}\selectfont}c}
    @{}
}
\toprule
\shortstack{Multi-Scale Harmonic\\Structural Encoding}
& \shortstack{Cross-Order Harmonic\\Chirality Representation}
& \shortstack{Diversity-Aware\\Triplet Selection}
& OpenEarthMap & Vaihingen & VDD \\
\midrule
\(\times\) & \(\checkmark\) & \(\checkmark\)
& 48.17 & 63.56 & 65.71 \\

\(\checkmark\) & \(\times\) & \(\checkmark\)
& 48.07 & 63.58 & 65.62 \\

\(\checkmark\) & \(\checkmark\) & \(\times\)
& 48.19 & 63.64 & 65.72 \\

\(\checkmark\) & \(\checkmark\) & \(\checkmark\)
& \textbf{49.45} & \textbf{64.30} & \textbf{66.21} \\
\bottomrule
\end{tabular*}

\end{table*}

Table~\ref{tab:mh_d4vs_ablation} further examines the contributions of individual components within MH-D4VS. MH-D4VS consists of three components including Multi-Scale Harmonic Structural Encoding, Cross-Order Harmonic Chirality Representation and Diversity-Aware Triplet Selection. The complete configuration achieves mIoU scores of 49.45\%, 64.30\%, and 66.21\% on OpenEarthMap, Vaihingen, and VDD, respectively. Removing Multi-Scale Harmonic Structural Encoding reduces these scores to 48.17\%, 63.56\%, and 65.71\%, respectively. Removing Cross-Order Harmonic Chirality Representation yields 48.07\%, 63.58\%, and 65.62\%, while removing Diversity-Aware Triplet Selection results in 48.19\%, 63.64\%, and 65.72\%.

All three simplified configurations consistently reduce performance. Multi-Scale Harmonic Structural Encoding captures complementary directional information across different spatial extents. Cross-Order Harmonic Chirality Representation provides additional chirality-based discrimination between rotations and reflections. Diversity-Aware Triplet Selection further constrains the geometric complementarity of the candidate set beyond the structural preferences of individual views. These results indicate that the performance of MH-D4VS arises from the combined contributions of multi-scale orientation descriptors, reflection-sensitive modeling, and set-level view selection, rather than from a single structural metric.

\subsubsection{Ablation of OAMR}
\label{subsubsec:oamr_ablation}

\begin{table*}[t]
\centering
\caption{Ablation study of different components in OAMR (mIoU, \%).}
\label{tab:oamr_ablation}

\fontsize{7}{7.2}\selectfont
\setlength{\tabcolsep}{2pt}
\renewcommand{\arraystretch}{1.15}

\begin{tabular*}{\textwidth}{
    @{\extracolsep{\fill}}
    *{7}{>{\fontsize{7}{7.2}\selectfont}c}
    @{}
}
\toprule
\shortstack{Adaptive Local\\Subspace Modeling}
& \shortstack{Refined Cross-View\\Metric-Guided Refit}
& \shortstack{Cross-Scale\\Guidance}
& OpenEarthMap & Vaihingen & VDD \\
\midrule
\(\times\) & \(\checkmark\) & \(\checkmark\)
& 44.13 & 60.88 & 64.35 \\

\(\checkmark\) & \(\times\) & \(\checkmark\)
& 44.24 & 60.90 & 64.36 \\

\(\checkmark\) & \(\checkmark\) & \(\times\)
& 46.26 & 63.74 & 65.64 \\

\(\checkmark\) & \(\checkmark\) & \(\checkmark\)
& \textbf{49.45} & \textbf{64.30} & \textbf{66.21} \\
\bottomrule
\end{tabular*}

\end{table*}

Table~\ref{tab:oamr_ablation} examines the roles of Adaptive Local Subspace Modeling, Refined Cross-View Metric-Guided Refit, and Cross-Scale Guidance in OAMR by removing one component at a time. The complete OAMR configuration achieves mIoU scores of 49.45\%, 64.30\%, and 66.21\% on OpenEarthMap, Vaihingen, and VDD, respectively. Removing any component reduces performance.

Without Adaptive Local Subspace Modeling, performance drops to 44.13\%, 60.88\%, and 64.35\%, indicating that a single fixed manifold reference cannot adequately characterize the feature distributions of different regions. Removing Refined Cross-View Metric-Guided Refit yields 44.24\%, 60.90\%, and 64.36\%, respectively. This suggests that initial consistency alone is insufficient for reliable cross-view assessment, whereas metric refinement based on local manifold geometry and neighborhood relationships further improves anchor and candidate-source estimation.

When only Cross-Scale Guidance is removed, performance decreases to 46.26\%, 63.74\%, and 65.64\%. The larger decline on OpenEarthMap indicates that fine-grained candidate-source evidence from the high-resolution guidance level can supplement local decisions at the target repair level. Overall, Adaptive Local Subspace Modeling provides the structural reference, Refined Cross-View Metric-Guided Refit improves cross-view reliability estimation, and Cross-Scale Guidance extends repair to additional regions with fine-grained support. Together, these components form the complete OAMR repair process.

\subsubsection{Visual Analysis of Feature Differences and Manifold Repair under \(\mathrm{D}_4\) Transformations}
\label{subsubsec:d4_feature_repair_visualization}

\begin{figure*}
	\centering
	\includegraphics[width=1.0\textwidth]{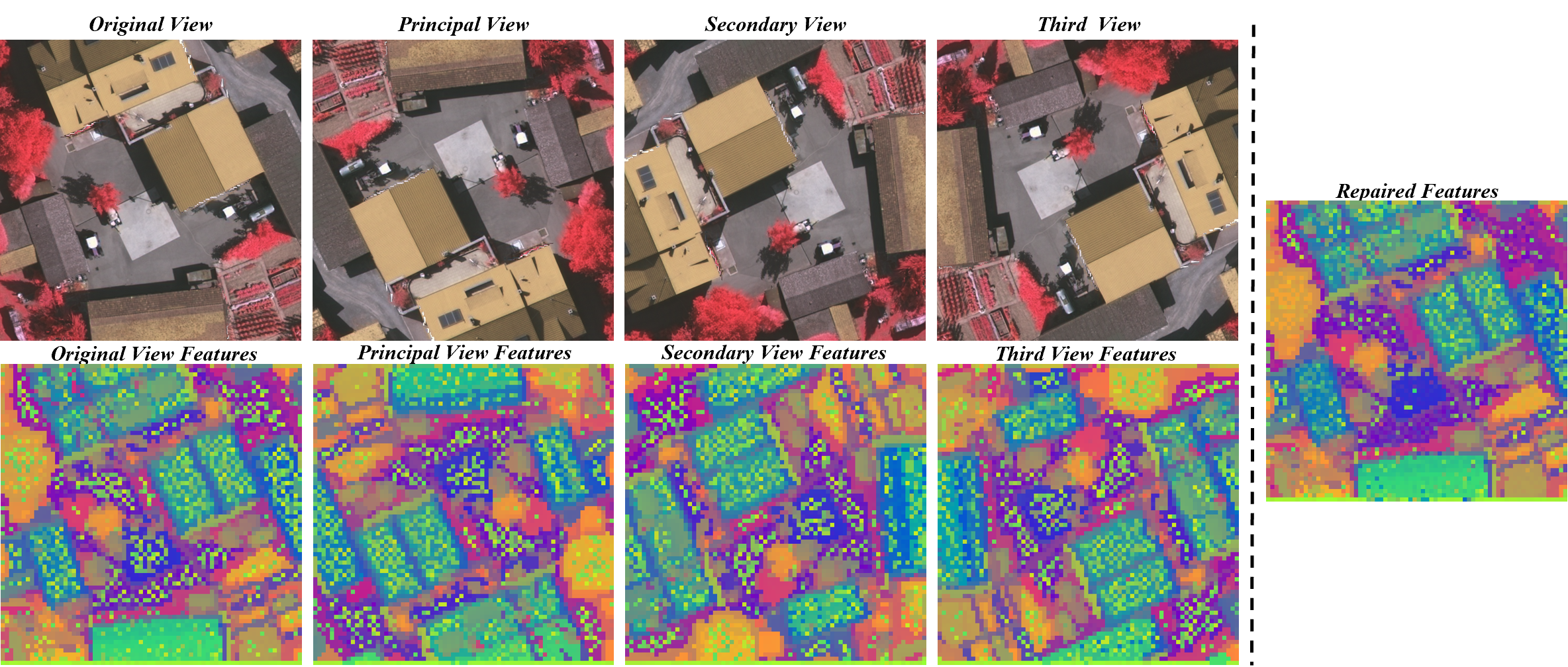}
	\caption{Visualization of multi-view feature repair with OAMR. On the left, the top row shows the Original View and three transformed views (Principal, Secondary, and Third), while the bottom row visualizes their features in a shared PCA space after alignment to the original coordinate system. The right panel shows the features repaired by OAMR. All features use the same PCA projection and color mapping. Colors indicate only relative feature projections in the low-dimensional PCA space and do not correspond to specific semantic classes.}
	\label{fig:oamr_feature_visualization}
\end{figure*}

Figure~\ref{fig:oamr_feature_visualization} presents feature visualizations before and after OAMR repair in a shared PCA space. The colors represent only the projections of high-dimensional features onto a low-dimensional principal component space and do not correspond to specific semantic classes. We therefore focus on the spatial distribution of different regions, local continuity, and relative changes before and after repair.

At the global level, the dominant feature distribution patterns remain largely consistent before and after OAMR repair. The locations and contours of large-scale regions, as well as their relative spatial relationships, show no substantial changes. This suggests that the repair process does not introduce extensive feature reconstruction and largely preserves the major spatial structures of the original features. Locally, the repaired feature representations exhibit greater continuity. Some isolated responses that previously differed markedly from their surroundings, along with fragmented local variations, become less pronounced. Visual transitions between adjacent tokens become more gradual, resulting in more spatially consistent local regions. Meanwhile, the major boundaries between existing regions remain distinct, with no evident widespread blurring or merging of regions. These observations are consistent with the selective repair mechanism of OAMR. 

\begin{figure*}
	\centering
	\includegraphics[width=1.0\textwidth]{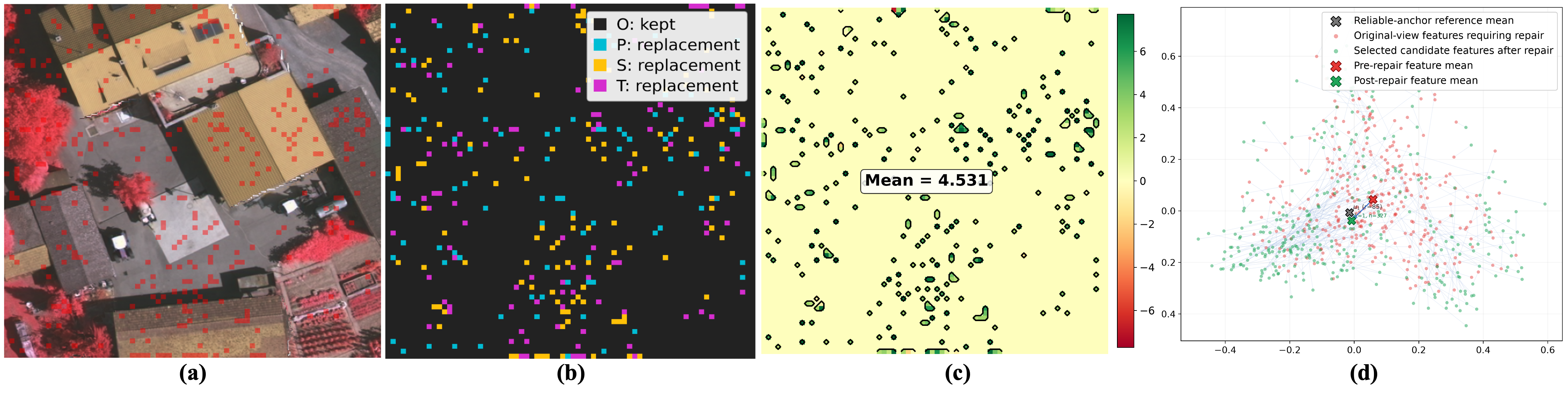}
	\caption{Spatial distribution, replacement sources, and changes in local manifold consistency for selective feature repair with OAMR. (a) Final feature repair regions, with red locations indicating feature replacement. (b) Distribution of final replacement sources. \(O\) indicates retention of the original-view features, while \(P\), \(S\), and \(T\) indicate replacement with aligned features from the Principal, Secondary, and Third candidate views, respectively. (c) Spatial distribution of improvements in local manifold anomaly scores. Colors represent the minimum local subspace anomaly score before repair minus the anomaly score after repair. Positive values indicate improved consistency with the local manifold, whereas negative values indicate increased deviation. Black contours delineate the final repair locations, and the displayed values report the mean improvement within the final repair regions. (d) Changes in the feature distribution at the final repair locations in a shared PCA space. Gray crosses denote the means of reliable anchor features. Red and green points represent the original-view features before repair and the candidate features used for replacement, respectively. Connecting lines indicate paired feature changes at the same spatial location, while red and green crosses denote the feature means before and after repair, respectively.}
	\label{fig:oamr_repair_analysis}
\end{figure*}

As shown in Figure~\ref{fig:oamr_repair_analysis}(a), feature repair in OAMR is distinctly selective. The replaced tokens are distributed across different spatial regions of the image, while other locations retain their original-view features. This agrees with the design of OAMR, which performs replacement at locations satisfying both the anomaly criteria and candidate acceptance conditions, indicating that the method does not extensively reconstruct the entire feature map.

Figure~\ref{fig:oamr_repair_analysis}(b) further illustrates the actual feature sources at the repaired locations. All three candidate views, Principal, Secondary, and Third, contribute to the final repair. Their contributions are spatially interspersed, with no single candidate serving as the replacement source for all locations. This indicates that OAMR independently evaluates multiple candidate features at each anomalous location and selects a candidate token that satisfies the local manifold constraints and better suits that location, rather than using a fixed transformed view as the global repair source.

Figure~\ref{fig:oamr_repair_analysis}(c) shows that anomaly-score changes remain close to zero at locations where no replacement occurs, whereas more pronounced changes are concentrated within the final repair regions delineated by black contours. This is because unrepaired locations retain the same original features before and after repair, leaving their local subspace anomaly scores essentially unchanged. Within the final repair regions, anomaly-score improvements are predominantly positive, with a mean improvement of 4.531. Since this metric is defined as the anomaly score before repair minus that after repair, positive values indicate that feature replacement reduces the standardized deviation of the corresponding token from its nearest local subspace. These results show that, for the current sample, the candidate features selected by OAMR generally reduce local manifold anomalies at the repaired locations.

Figure~\ref{fig:oamr_repair_analysis}(d) further illustrates the feature changes of these repaired tokens in a shared PCA projection space. Features at different locations move in different directions after repair, rather than undergoing a uniform translation. This observation agrees with the location-specific candidate selection mechanism of OAMR based on local structure. The mean of the repaired features moves closer to the reference mean of the reliable anchors. Meanwhile, some features initially distant from the reference region move toward the reliable feature distribution after replacement. These observations are consistent with the reduction in local subspace anomaly scores shown in Figure~\ref{fig:oamr_repair_analysis}(c), providing intuitive qualitative evidence of improved feature consistency from a two-dimensional projection perspective.

\section{Conclusion}
\label{sec:conclusion}

This work proposes GFMR to address the underutilization of geometric information, inconsistent local representations across views, and distribution shifts between samples in SAM3-based OVSS of remote sensing images. Without modifying the core architecture or original prediction pathway of SAM3, GFMR integrates geometry-aware view selection, original-view-anchored manifold repair, and online PixelDecoder adaptation at test time into a unified inference pipeline. Specifically, MH-D4VS uses multi-scale complex harmonics, chirality, and set-level geometric constraints to select three representative and complementary \(\mathrm{D}_4\) candidate views. Anchored to the original view, OAMR uses adaptive local subspaces, refined cross-view metrics, and cross-scale guidance to repair locally anomalous representations using only candidate features with reliable support. PD-TTA uses local--global semantic entropy computed from the outputs of the frozen semantic head as a self-supervised signal to update the affine parameters of two GroupNorm layers in the PixelDecoder online during image segmentation. MH-D4VS and OAMR require no gradient updates, while PD-TTA requires neither target-dataset-level training nor additional annotations.

Experiments on eight remote sensing benchmarks demonstrate that GFMR achieves an average mIoU of 55.6\%, with consistent gains across satellite, aerial, and UAV imagery. Transferring MH-D4VS and OAMR to another SAM3-based inference framework, ProC-SAM3, also yields substantial segmentation improvements, demonstrating the compatibility of the proposed modules with different frameworks. Ablation studies further confirm the complementarity of the three components: the training-free MH-D4VS and OAMR modules account for the primary performance gains, while PD-TTA provides consistent additional improvements at the sample level. There remains considerable room to improve the effectiveness of test-time adaptation (TTA) for SAM3 inference. Future work will further investigate the integration of TTA strategies into the SAM3 inference process.

\bibliographystyle{cas-model2-names}

\bibliography{GFMR}

\end{document}